\documentclass[conference]{IEEEtran}
\IEEEoverridecommandlockouts
\usepackage{cite}
\usepackage{amsmath,amssymb,amsfonts}
\usepackage{graphicx}
\usepackage{textcomp}
\usepackage{xcolor}

\usepackage{xcolor} 
\usepackage[most]{tcolorbox}

\usepackage{hyperref}
\usepackage{cleveref}

\usepackage{listings}
\definecolor{codegreen}{rgb}{0,0.6,0}
\definecolor{codegray}{rgb}{0.5,0.5,0.5}
\definecolor{codepurple}{rgb}{0.58,0,0.82}
\definecolor{backcolour}{rgb}{0.96,0.96,0.96}

\lstdefinestyle{scmstyle}{
    backgroundcolor=\color{backcolour},   
    commentstyle=\color{codegreen},
    keywordstyle=\color{magenta},
    numberstyle=\tiny\color{codegray},
    stringstyle=\color{codepurple},
    basicstyle=\ttfamily\footnotesize, 
    breakatwhitespace=false,         
    breaklines=true,                 
    captionpos=b,                    
    keepspaces=true,                 
    numbers=left,                    
    numbersep=5pt,                  
    showspaces=false,                
    showstringspaces=false,
    showtabs=false,                  
    tabsize=2,
    frame=single,                    
    rulecolor=\color{gray!30}
}

\usepackage{fancybox,framed}    

\usepackage{enumitem} 

\usepackage{booktabs} 
\usepackage{xfp}

\newcommand{\gameonewinrate}{32.98} 
\newcommand{\gametwowinrate}{23.88} 
\newcommand{\gamethreewinrate}{17.23} 
\newcommand{\levelonewinrate}{27.24} 
\newcommand{\levelthreewinrate}{18.91} 
\newcommand{\levelfourwinrate}{12.18} 
\newcommand{\diffthinkgameone}{50.70} 
\newcommand{\diffthinkgametwo}{27.57} 
\newcommand{\diffthinkgamethree}{27.30} 
\newcommand{\diffthinkoverall}{35.51} 
\newcommand{\diffcausalwin}{5.35} 
\newcommand{\diffcausaltime}{17} 

\def\BibTeX{{\rm B\kern-.05em{\sc i\kern-.025em b}\kern-.08em
    T\kern-.1667em\lower.7ex\hbox{E}\kern-.125emX}}

\usepackage{silence}
\usepackage{microtype}

\title{Spatial Reasoning in LLM Game Agents: Impact of Causal Context and Multi-Step Planning
}

\makeatletter
\newcommand{\linebreakand}{%
  \end{@IEEEauthorhalign}
  \hfill\mbox{}\par
  \mbox{}\hfill\begin{@IEEEauthorhalign}
}
\makeatother

\author{
  \IEEEauthorblockN{Mohit Jiwatode} 
  \IEEEauthorblockA{\textit{Institute for Information Processing}\\
  \textit{Leibniz University Hannover}\\
  Hannover, Germany \\
  jiwatode@tnt.uni-hannover.de}
  \and
  \IEEEauthorblockN{Ronja Fuchs} 
  \IEEEauthorblockA{\textit{Institute for Information Processing}\\
  \textit{Leibniz University Hannover}\\
  Hannover, Germany \\
  fuchsron@tnt.uni-hannover.de}
  \and
  \IEEEauthorblockN{Robin Schmöcker} 
  \IEEEauthorblockA{\textit{Institute for Information Processing}\\
  \textit{Leibniz University Hannover}\\
  Hannover, Germany \\
  schmoecker@tnt.uni-hannover.de}
  \linebreakand
  \IEEEauthorblockN{Bodo Rosenhahn} 
  \IEEEauthorblockA{\textit{Institute for Information Processing}\\
  \textit{Leibniz University Hannover}\\
  Hannover, Germany \\
  rosenhahn@tnt.uni-hannover.de}
  \and
  \IEEEauthorblockN{Alexander Dockhorn} 
  \IEEEauthorblockA{\textit{SDU Metaverse Lab}\\
  \textit{University of Southern Denmark}\\
  Odense, Denmark \\
  adoc@sdu.dk}
}

\begin{document}

\IEEEoverridecommandlockouts
\IEEEpubid{\makebox[\columnwidth]{ 979-8-3315-9476-3/26/\$31.00 \copyright2026 IEEE\hfill} 
\hspace{\columnsep}\makebox[\columnwidth]{ }}

\maketitle

\IEEEpubidadjcol

\begin{abstract}
LLM-based game agents often perform poorly on more complex tasks. This work examines whether these failures are linked to limited spatial reasoning and evaluates whether causal prompt augmentation and multi-step planning can improve win-rates while managing response latency.
Using the open-source Qwen3 model family, we conduct experiments across varying model scales, reasoning modes, and planning horizons. 
We further introduce a focused GVGAI benchmark consisting of three custom games with five difficulty levels to isolate spatial navigation. 
The evaluation follows two paradigms: an initial ``positioning experiment'' to test an agent's ability to find its exact coordinates, and a study of game-play success. 
Our results show that while larger models with an enabled thinking mode identify their positions more accurately, overall performance in coordinate matching remains limited for smaller models.
Win rates decrease as game levels and layout complexity increase, validating the benchmark's difficulty scaling. 
Integrating causal context into the prompts tends to improve the agents' success rates, particularly for bigger models. 
While enabling thinking mode and longer planning horizons significantly improve performance, multi-step planning further reduces mean per-step response times, offering a practical trade-off between reasoning depth and execution speed.

\end{abstract}



\section{Introduction}

General game playing is a useful stress test for Large Language Model (LLM) agents, since it requires instruction following, state tracking, sequential control, and adaptation to unfamiliar rules~\cite{li2025gvgai_llm,eberhardinger2025code}.
Especially in a language-based setting, such as the General Video Game AI (GVGAI)~\cite{liebana2022general} framework, LLM agents can be tested well.  
GVGAI specifies games in the Video Game Description Language (VGDL), which exposes mechanics in symbolic form while still requiring step-by-step decision-making across diverse environments~\cite{liebana2022general,schaul2014extensible,torrado2018gvgai_gym,perezliebana2019gvgai}. 

A model that performs well in GVGAI must interpret structured observations, ground actions correctly, update behavior after each transition, and may even plan multiple steps ahead. This makes GVGAI relevant to recent work on grounded spatial reasoning and sequential instruction following, where performance depends on agent positioning, pathing, and coherent action selection with short-horizon intent over time~\cite{janner2018representation,rizvi2024sparc,chen2024sifo}.
However, recent work suggests that strong LLMs still perform poorly on these game-playing tasks~\cite{li2025gvgai_llm}. While typical problems like large state or action spaces, long planning horizons, or limited learning due to sparse rewards remain, LLMs also face model-specific problems such as limited spatial reasoning capabilities, slow response times, and incongruent actions~\cite{li2025gvgai_llm}.

Rather than asking whether a model can solve the full GVGAI suite, we investigate common shortcomings of LLM agents and potential ways to overcome them. One motivation for our study is causal prompt augmentation, which proposes that LLMs benefit from explicit structural descriptions of environment dynamics and how they should be interpreted~\cite{jiwatode2026trace2mechanics,chen2025causal,pan2025bayesian}. Following this line of work, we compare the performance of LLM-based agents with and without causal prompt augmentation to assess whether it improves spatial reasoning in the environment. 
We further consider the agent's thinking mode as another factor that may shape spatial reasoning. While enabling thinking mode often improves performance and response quality, it also substantially increases response time~\cite{aggarwal2025optimalthinkingbench}. To address this trade-off, we study multi-step planning and its interaction with causal augmentation and thinking mode as a means of mitigating the cumulative latency incurred by single-step action selection. Supplementary materials, including the appendix and the source code repository for experiment reproducibility, are available under:  \url{https://drive.google.com/drive/folders/1c3iPa8RbW0xTOuBAkSr4AA8yvlsPM2eR?usp=sharing}.

To evaluate the LLM agent's performance, we introduce a controlled spatial benchmark inside GVGAI with three custom games and five difficulty levels per game. For each of these, we test the LLM in various configurations to evaluate the impact of causal context, thinking mode, and planning horizon on its spatial reasoning capabilities and overall success.

Our contributions are:
\begin{itemize}
    \item We introduce a focused GVGAI benchmark consisting of three custom games across five difficulty levels, designed specifically to isolate symbolic navigation and geometric reasoning from complex game mechanics.
    \item We quantify the impact of providing agents with structured environment dynamics in the form of a causal model on the win-rate, completion rates, and response latency.
    \item We analyze how the thinking mode and planning horizon change both the success and the response time of tested LLM-based agents.
    \item We provide an initial evaluation of how the Qwen3 model family (ranging from 0.6B to 8B parameters) handles symbolic tasks, highlighting the performance gap between model sizes and the challenges smaller models face with self-localization and 2D navigation.
\end{itemize}
The remainder of this paper is organized as follows. \Cref{sec:foundations} introduces the necessary foundations and \Cref{sec:related} situates the work within the existing literature. \Cref{sec:methodology} outlines the experimental design and methodology, followed by a presentation of results in \Cref{sec:results} and their discussion in \Cref{sec:discussion}. Limitations are discussed in \Cref{sec:limitations}. Finally, \Cref{sec:conclusion} summarizes the contributions and identifies directions for future work.

\section{Foundations}
\label{sec:foundations}

\subsection{Video Game Description Language (VGDL) and the General Video Game AI (GVGAI) framework} 

VGDL~\cite{schaul2014extensible} is a high-level language for specifying 2D games. A VGDL game definition comprises both the game dynamics and one or more level layouts. Because games can be described compactly in VGDL, it is well-suited for the controlled environments used in this paper. We use VGDL to define these environments and to construct a symbolic observation space for the agent. This observation space includes a description of the game mechanics as well as the current state represented as a matrix of object names. In contrast to raw pixel arrays, such a symbolic representation is directly interpretable by LLMs. The concrete VGDL specifications of the games used in this work are provided in \Cref{sec:games}. While VGDL is used to describe the games, we rely on the GVGAI framework~\cite{liebana2022general} to execute them and to evaluate agent performance.

\subsection{Problem formalization} The interaction between a player and a single-player game defined in VGDL can be modeled as a sequential decision process. At time step $i=0$, an initial state $s_0$ is selected from one of the predefined level layouts. At each subsequent step $i$, the player receives an observation $o_i$ of the current state $s_i$, chooses a legal action $a_i$, and the environment transitions to the next state $s_{i+1}$ while producing a reward $r_i$. This process continues until a terminal condition is reached at time step $N$. The objective is to maximize the cumulative reward $\sum_{i=0}^{N-1} r_i$ and, where applicable, satisfy the game's win condition. In this paper, the player in this interaction loop is an LLM-based agent. The prompting scheme used to interface with the LLM is described in \Cref{sec:prompts}.

\section{Related Work}\label{sec:related}

The GVGAI framework is an established benchmark for general game-playing, as it consists of a diverse set of more than 100 games. Broader game benchmarks such as SmartPlay~\cite{wu2023smartplay}, GameBench~\cite{allen2024gamebench}, BALROG~\cite{paglieri2024balrog}, and gg-bench~\cite{zhu2025ggbench} show that the difficulties observed in GVGAI are part of a wider pattern in LLM game reasoning. They emphasize strategic reasoning, long-horizon control, and generalization across game families. Our work does not attempt to compete with those suites as a broad benchmark. Instead, it uses a small, controlled GVGAI benchmark to test whether spatial reasoning and latency can be examined more cleanly than in a full sweep.

A large body of research has been done concerning the usage of LLMs as the decision-maker in sequential decision-making problems. For example, Yao et al.~\cite{yao2023react} consider a fact verification task where the agent has to interact with the Wikipedia API. Park et al.~\cite{park2023generative} employ LLMs in a multi-agent sandbox environment which resembles \textit{The Sims}\footnote{\url{https://www.ea.com/games/the-sims}; Last accessed \today}, Wang et al.~\cite{wang2023voyager} showed that an LLM-based agent can accumulate skills over long trajectories in an open-ended game world, and Eberhardinger et al. ~\cite{eberhardinger2025code} used LLMs for code generation, which was reframed as a sequential decision-making task.

The causal augmentation used in this paper is motivated by work that treats environment dynamics as an explicit intermediate representation rather than leaving all reasoning implicit in the prompt-response loop. Recent work on mechanic induction from gameplay traces argues that LLMs can infer game rules from transitions and use those rules as structured knowledge~\cite{jiwatode2026trace2mechanics}. Related work on causal-aware agents also suggests that explicit structure can improve decision-making in sequential environments ~\cite{yu2024adam,chen2025causal,pan2025bayesian,lampinen2023passive,zhao2025curious}. In contrast, our work does not learn a full causal model online. Instead, it asks whether adding explicit causal context measurably improves performance in spatial GVGAI tasks.

\section{Methodology}
\label{sec:methodology}

Li et al. \cite{li2025gvgai_llm} showed that the then state-of-the-art models struggled to solve games in the GVGAI framework. The two main issues are low win-rates and slow response times. It is important to understand why these problems occur and how to improve performance in these areas.
In our work, we aim to quantify the ability of LLMs to solve symbolic spatial tasks and find out if causal context, thinking mode, and multi-step planning influence the overall performance of the model. In order to ensure consistency, we used the Goal-Question-Metric method as defined by Wohlin et al.~\cite{wohlin2012experimentation} and formulated the following three research questions (RQs):

\begin{framed}
    \begin{enumerate}
        \item[\textbf{RQ1}] To what extent can an LLM-based agent solve simplified symbolic spatial tasks, and how does performance degrade with increasing layout difficulty?
        \item[\textbf{RQ2}] Does added causal context improve spatial-control performance?
        \item[\textbf{RQ3}] How do causal augmentation, thinking mode, and multi-step planning affect the trade-off between success and response latency? 
    \end{enumerate}
\end{framed}

\begin{figure*}[t]
  \centering
  \includegraphics[width=0.65\textwidth]{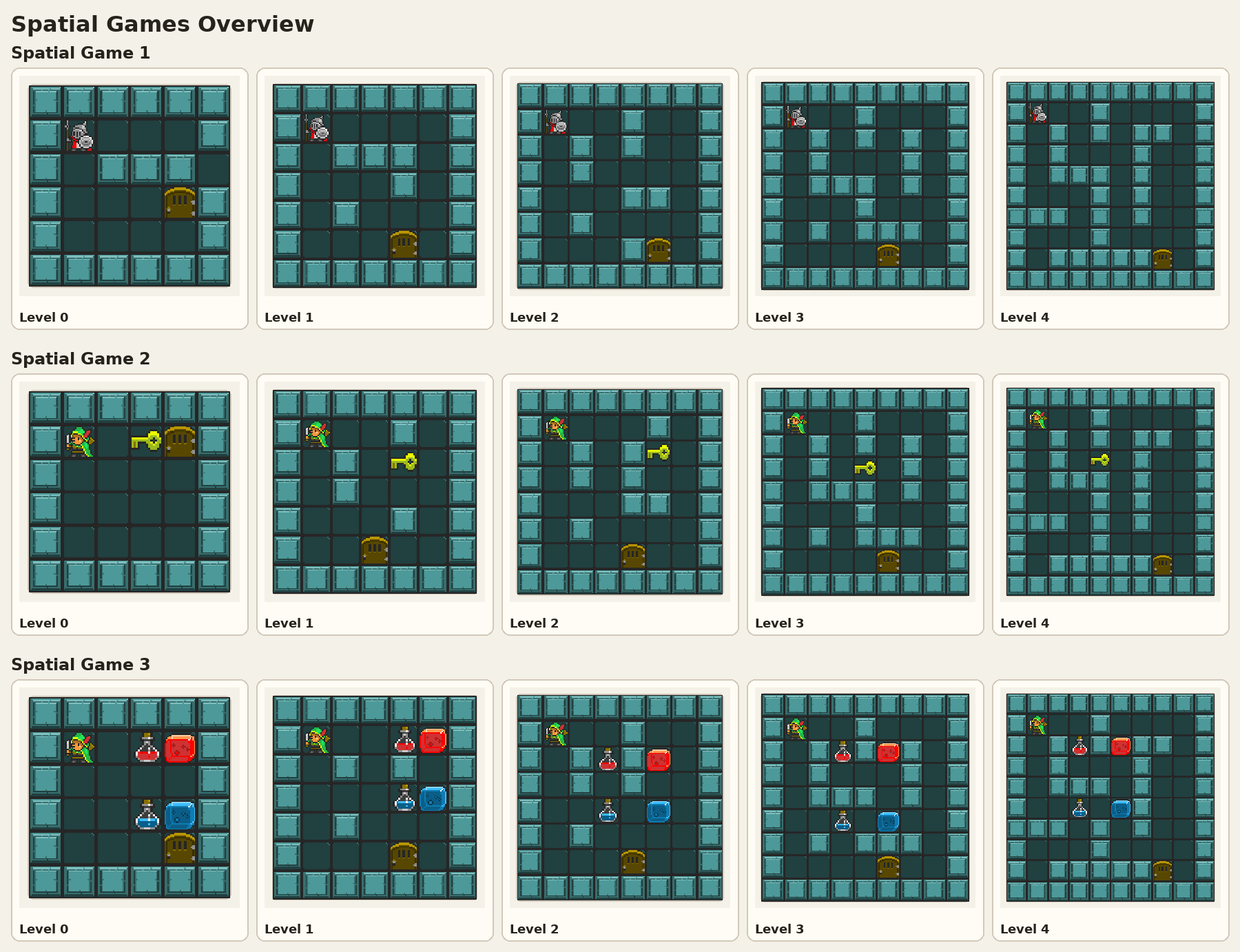}
  \caption{Overview of all the levels in the spatial games.}
  \label{fig:spatial_games_overview}
\end{figure*}

\textbf{RQ1} quantifies the intrinsic spatial reasoning capabilities of LLMs within a symbolic environment. By focusing on tasks where the agent's primary objectives are self-localization and path-finding toward a goal tile, we isolate geometric reasoning from complex game mechanics. This allows us to determine if an LLM can navigate based solely on the grid's topology or if performance significantly degrades as layout complexity, such as grid size and the density of immovable walls, increases.

\textbf{RQ2} evaluates whether providing the agent with structured game mechanics in the form of a causal model improves performance and reduces reasoning latency. We investigate whether an explicit causal structure negates the need for the LLM to model the environment dynamics during its internal reasoning process. By augmenting prompts with these structured descriptions, we determine if grounded logical context assists the LLM in interpreting symbolic observations and transition rules more efficiently than prose alone. This analysis clarifies whether spatial reasoning deficiencies stem from inherent reasoning limitations or from the lack of a structured framework to process environmental causal effects.

\textbf{RQ3} tests whether multi-step planning improves win rates and reduces the mean time per step. Long-term planning evaluates extended spatial reasoning by requiring the agent to validate path reachability and maintain trajectory consistency within a single context window. This approach mitigates "spatial drift" or the accumulation of local positioning errors during single-step execution by forcing the model to pre-calculate a coherent sequence of movements. Furthermore, we analyze the trade-off between the high response time of ``thinking'' modes and the efficiency gains of planning horizons to determine if increased depth reduces the average computational cost per action without compromising navigation success.

To address these questions, we use three sets of metrics across three dimensions, i.e., the levels per game, the average per game, and the average across the whole benchmark:
\begin{itemize}[leftmargin=3.2em, labelsep=1em, itemsep=0em]
    \item[\textbf{M1.1}] \textbf{Completion Rate:} The proportion of runs that reached any termination condition (win, loss, or timeout).
    \item[\textbf{M1.2}] \textbf{Win-Rate:} The proportion of runs that successfully reached the winning state.
    \item[\textbf{M1.3}] \textbf{Mean Time per Step:} The total execution time of a run divided by the number of steps taken.
    \smallskip 
    \item[\textbf{M2.1}] \textbf{Causal Impact:} The performance delta in metrics \textbf{M1.1--M1.3} between agents with and without causal prompt augmentation.
    \smallskip
    \item[\textbf{M3.1}] \textbf{Reasoning Impact:} The performance delta in metrics \textbf{M1.1--M1.3} between agents using thinking mode versus standard inference.
    \item[\textbf{M3.2}] \textbf{Planning Impact:} The performance delta in metrics \textbf{M1.1--M1.3} across planning horizons of \mbox{$H \in \{1, 5, 10\}$}.
\end{itemize}

\subsection{Benchmark Design}
\vspace{0.5em}

The study utilizes three games focusing on spatial reasoning, called \texttt{spatial game 1}, \texttt{2}, and \texttt{3}, representing games of increasing logical complexity. Each environment is defined in VGDL~\cite{schaul2014extensible} and simulated using the GVGAI framework~\cite{liebana2022general}. In addition to standard win/loss termination conditions, each game incorporates a fixed temporal limit (timeout) as a termination condition.

\subsection{Game Taxonomy}
\label{sec:games}
\vspace{0.5em}
\begin{itemize}
    \item \texttt{Spatial Game 1:} A fundamental path-planning task requiring the agent to navigate a static maze to reach the exit door in a dungeon.
    \item \texttt{Spatial Game 2:} Copies the previous game mechanics and further introduces a sequential dependency. The agent must first acquire a key to unlock the exit. The challenge lies in the presence of a visible but locked exit.
    \item \texttt{Spatial Game 3:} Copies the previous game mechanics and introduces a more complex multi-step task. The agent can collect a red and a blue key to open their respective doors before the final goal. Each opened door and obtained key increases the agent's score.
\end{itemize}

Each game consists of five levels (L0-L4) with increasing difficulty, controlled through grid size expansion (from $6 \times 6$ to $10 \times 10$) and the introduction of dense, maze-like structures and bottlenecks that necessitate advanced spatial reasoning. The games and the corresponding levels are shown in \Cref{fig:spatial_games_overview}\footnote{The VGDL files for the games can be found in the supplementary material.}.

\subsection{Experiment Setup}
\begin{figure}[t]
  \centering
  \includegraphics[width=0.8\columnwidth]{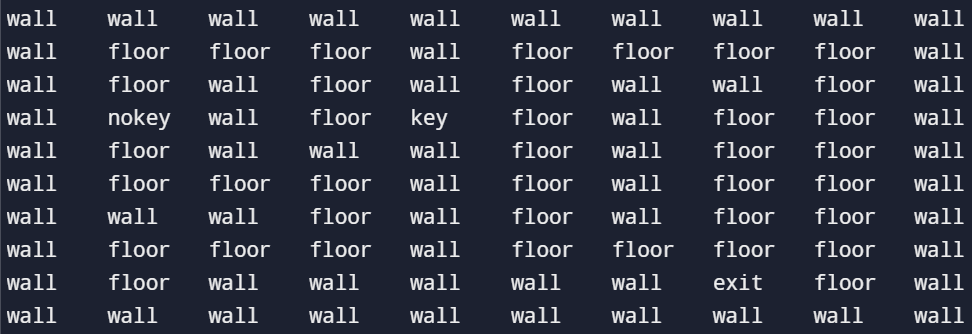}
  \caption{An example of the current state provided to the LLM as an input.}
  \label{fig:state_example}
\end{figure}

\vspace{0.5em}
Leveraging the GVGAI framework~\cite{liebana2022general,torrado2018gvgai_gym}, game states are extracted as ASCII symbol-based grids. To render these observations compatible with LLM inputs, we translate the raw symbols into their corresponding natural language descriptors using the game-specific \texttt{LevelMapping}. This semantic representation (shown in \Cref{fig:state_example}) serves as a part of the input to the LLM, providing a structured overview of the current game state. 

Agents interact with the environment through a set of five available actions: four directional movements (Up, Down, Left, Right) and a static Nil action, which maintains the avatar's current position. The rewards for the games are as follows: In Spatial Game 1, agents receive +1 for reaching the goal. Spatial Game 2 provides +1 for collecting the key and +2 for reaching the exit while holding it. Spatial Game 3 awards +1 for each colored key collected, +1 for each door opened, and +3 for reaching the final goal. Episodes terminate upon goal acquisition, avatar expiration, or the exhaustion of the allocated step and time limits.

\begin{figure}[t]
  \centering
  \includegraphics[width=0.7\columnwidth]{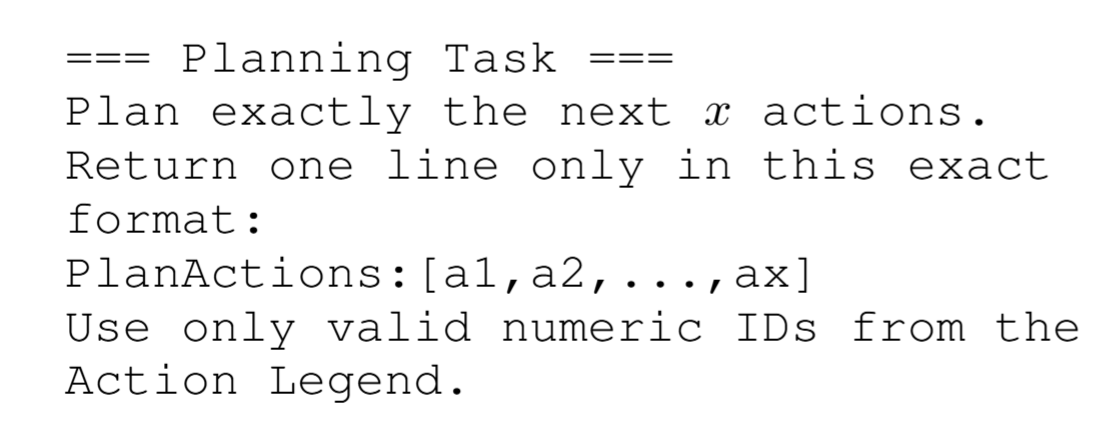}
  \caption{Prompt used to obtain next ``x'' actions from the LLMs }
  \label{fig:planning_prompt}
\end{figure}

Unlike the proprietary models used by Li et al.~\cite{li2025gvgai_llm}, we wanted to opt for open-source alternatives to ensure reproducibility of our experimental framework. Prior research~\cite{jiwatode2026trace2mechanics} has shown that the Qwen3~\cite{qwen3technicalreport} family of models shows good performance in understanding VGDL code and observations within the GVGAI games, and therefore, was chosen for our experiments. We specifically selected the 8B, 4B, 1.7B, and 0.6B parameter variants to ensure the experiments remain feasible on consumer-grade hardware. The model hyperparameters were configured according to the official documentation page to ensure optimal response quality~\cite{huggingfaceQwenQwen38BHugging}.


\subsection{Prompts}
\label{sec:prompts}
\vspace{0.5em}
The input prompt to the LLM agent consists of the following:
\begin{enumerate}
    \item A natural language description of the game generated by the LLM from the VGDL file of the game, adopted from ~\cite{li2025gvgai_llm}\footnote{The prompt used for generation, along with an example of the generated description, can be found in the supplementary material.}.
    \item Optionally, the prompt is augmented with a causal model obtained by the LLM using a strict blueprint adopted from Jiwatode et al.~\cite{jiwatode2026trace2mechanics}\footnote{The prompt used for generation of the causal model is in \Cref{fig:causal_model_prompt}, and the SCM blueprint can be found in the supplementary material.}.
    \item A list of all the available actions, along with their meanings obtained from the game.
    \item The current state of the game with the ASCII-based grid, with the symbols translated into the words using the level mappings from the VGDL. An example of a state is shown in \Cref{fig:state_example}.
    \item Optionally, in case of planning tasks, a prompt requiring the LLM to plan the next $x$ actions; this step is specifically designed to enforce a concrete output format, which is detailed in \Cref{fig:planning_prompt}.
    \item Optionally, in case of a self-localization task, a prompt requiring the LLM to find its own position on the grid in a strict format and return it along with the action feedback.
\end{enumerate}

\begin{figure}[t]
  \centering
  \includegraphics[width=0.65\columnwidth]{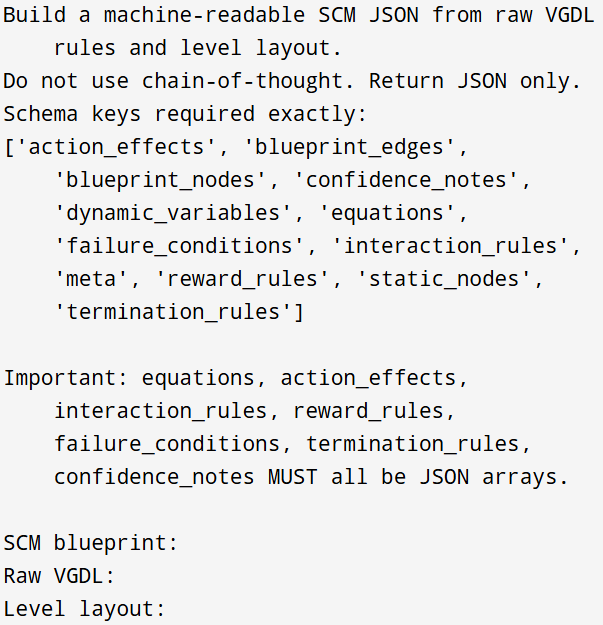}
  \caption{Prompt used to generate causal model from VGDL, Level layout, and SCM blueprint}
  \label{fig:causal_model_prompt}
\end{figure}

\subsection{Experiment 0 - Self-Localization Capability}\label{sec:exp_0}
\vspace{0.5em}
We conducted a preliminary experiment to determine if the LLM agent can accurately interpret the state representation by finding the avatar and returning its coordinates in a structured format. We utilized the three custom spatial games across all five difficulty levels, yielding 15 base game-level combinations. For each combination, we generated 10 unique test cases by taking the original layout and randomly relocating the avatar to a random empty tile. We tested the four Qwen3 model sizes (0.6B, 1.7B, 4B, and 8B) under both standard and thinking modes. To isolate this capability, we used a planning horizon of 1 without causal augmentation. The prompt was augmented to require the LLM to output the current coordinates of the avatar on the grid alongside its single next action. This resulted in 300 total queries per model (3 games × 5 levels × 10 cases × 2 thinking modes).

\subsection{Experiment 1 - Causal augmentation, Thinking, and Planning Impact Study}
\vspace{0.5em}
The experimental framework to study the impact of the model's internal reasoning, causal prompt augmentation and planning, follows a combinatorial design across four key dimensions: reasoning mode (thinking vs. non-thinking), model scale (8B, 4B, 1.7B, and 0.6B), causal augmentation (enabled vs. disabled), and planning horizon ($H \in \{1, 5, 10\}$). To account for the stochastic nature of the models induced by the non-zero temperature settings ($T=0.7$ for standard and $T=0.6$ for thinking mode), we conducted two independent trials for every unique parameter combination.

\begin{figure}[t]
  \centering
  \includegraphics[width=\columnwidth,]{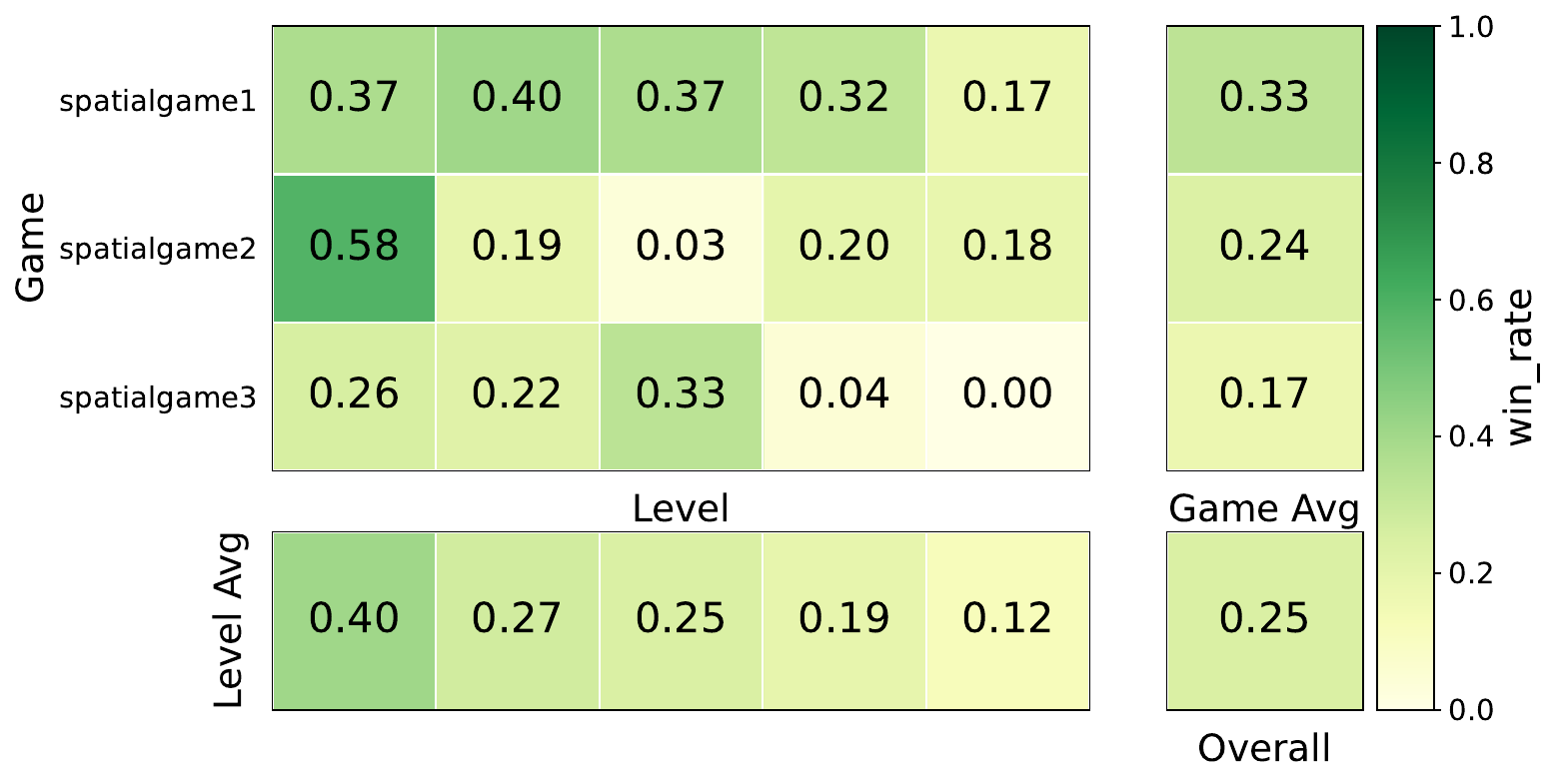}
  \caption{Win-rate heatmap for the spatial benchmark over games and levels.}
  \label{fig:spatial_winrate_heatmap}
\end{figure}

\begin{figure}[t]
  \centering
    \includegraphics[width=0.85\columnwidth]{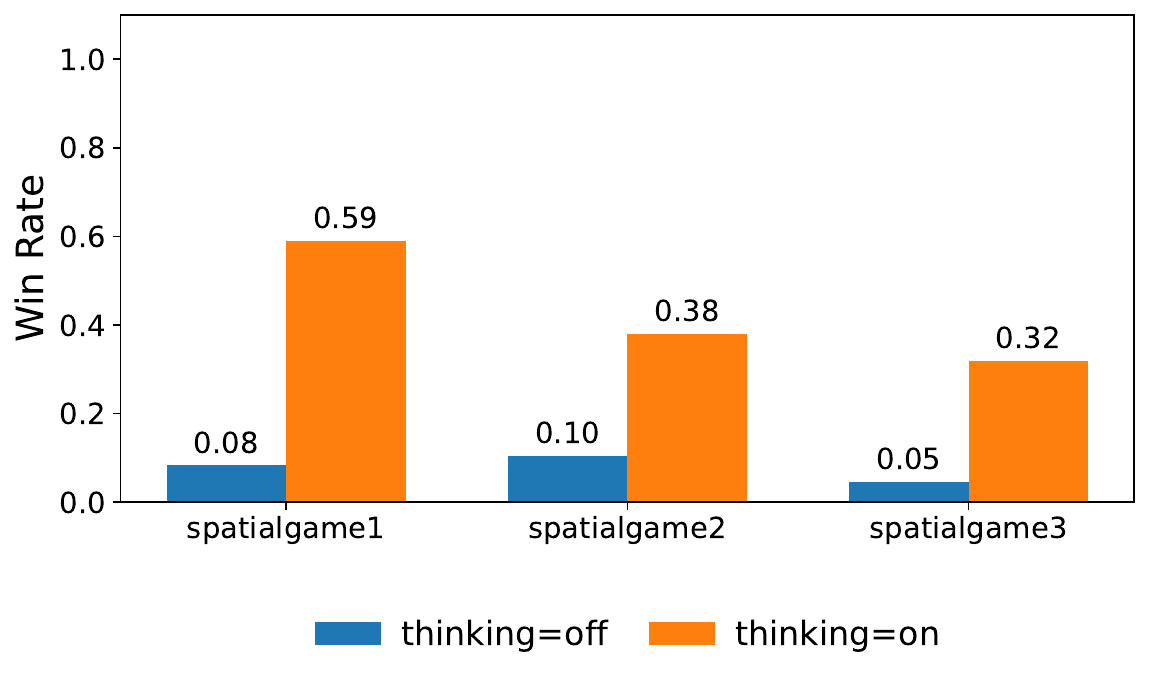}
  \caption{Win rate by game, split by thinking mode.}
  \label{fig:spatial_win_rate_by_game_thinking}
\end{figure}

\section{Results}
\label{sec:results}
\subsection{General Benchmark Data}
\Cref{fig:spatial_winrate_heatmap} shows the heatmap for the spatial benchmark over games and levels for all the models combined.
The agent performs best in \texttt{spatialgame1} (Win-rate $\approx$~\fpeval{round(\gameonewinrate,0)}\%), while it performed worse in \texttt{spatialgame2} ($\approx$~~\fpeval{round(\gametwowinrate,0)}\%) and \texttt{spatialgame3} ($\approx$~~\fpeval{round(\gamethreewinrate,0)}\%). 
For levels, the best results were achieved in levels with lower complexity (level 0: $\approx$~40\%, level 1: $\approx$~\fpeval{round(\levelonewinrate,0)}\%, level 2: $\approx$~25\%). 
In higher levels, the agent achieved a lower win-rate (level 3: $\approx$~\fpeval{round(\levelthreewinrate,0)}\%, level 4: $\approx$~\fpeval{round(\levelfourwinrate,0)}\%), suggesting that as the levels become more complex, the agent also struggles more.


The win-rate for each game with regard to thinking modes is shown in \Cref{fig:spatial_win_rate_by_game_thinking}. Overall, the model performed best with an enabled thinking mode. The difference in the win rates for the three games (1, 2, 3) was $\approx$~\fpeval{round(\diffthinkgameone,0)}\%, $\approx$~\fpeval{round(\diffthinkgametwo,0)}\%, and $\approx$~\fpeval{round(\diffthinkgamethree,0)}\%, respectively. Overall, the thinking or reasoning mode of the models improved the win-rates by $\approx$~\fpeval{round(\diffthinkoverall,0)}\% (see \Cref{tab:spatial_mode_summary}).

\vspace{0.1cm}
\subsection{Self localization}

LLM localization was poor. Performance scaled with model size; the smallest model failed entirely. In larger models, exact match rates increased with size despite similar Euclidean errors. Thinking mode reduced Euclidean error but showed no consistent pattern for exact matches.

\begin{figure}[t]
    \centering
    \includegraphics[width=0.8\linewidth]{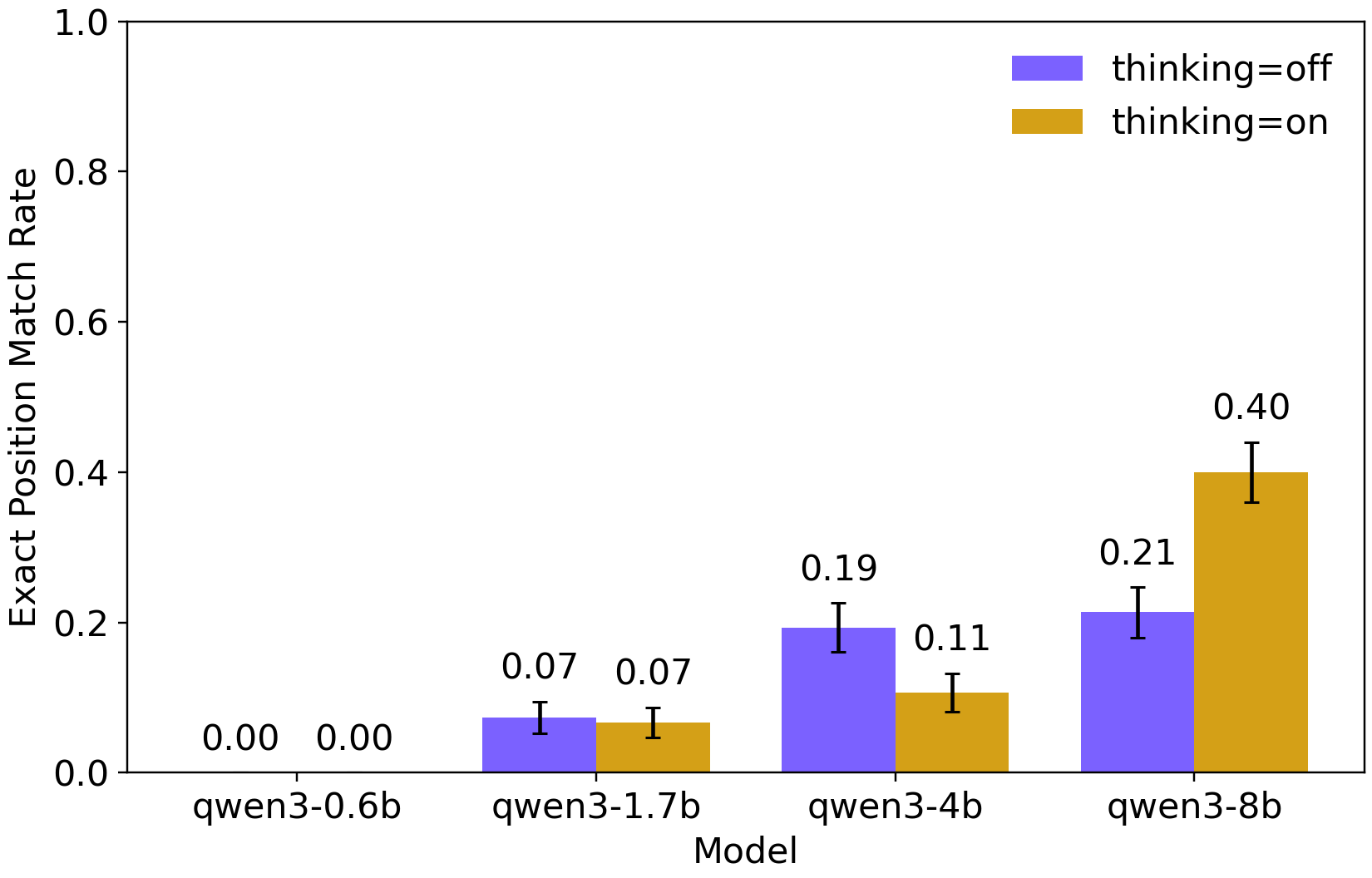}
    \caption{Model-wise Exact Position Match Rate by Thinking Mode. The 8B model shows a significant increase in exact matches when the thinking mode is enabled, while smaller models struggle to output the exact coordinates.}
    \label{fig:exact_position}
\end{figure}

\begin{figure}[t]
    \centering
    \includegraphics[width=0.8\linewidth]{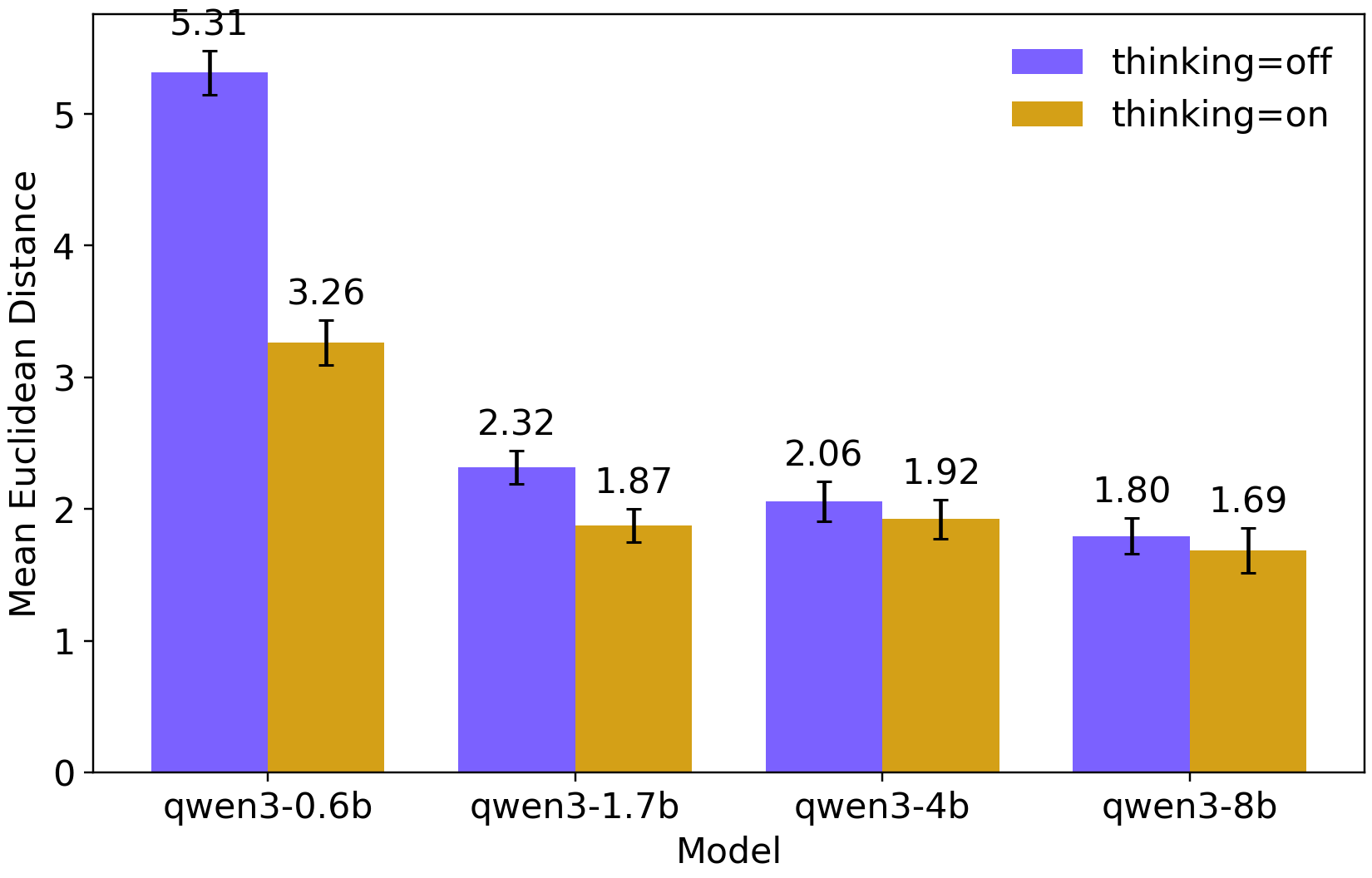}
    \caption{Model-wise Position Error by Thinking Mode. Enabling thinking mode consistently reduces the Mean Euclidean Distance between the predicted and actual coordinates across all evaluated model scales.}
    \label{fig:euclidean_distance}
\end{figure}

\subsection{Spatial reasoning: causal prompting, thinking, and planning}
\begin{figure}
    \centering
    \includegraphics[width=0.9\columnwidth]{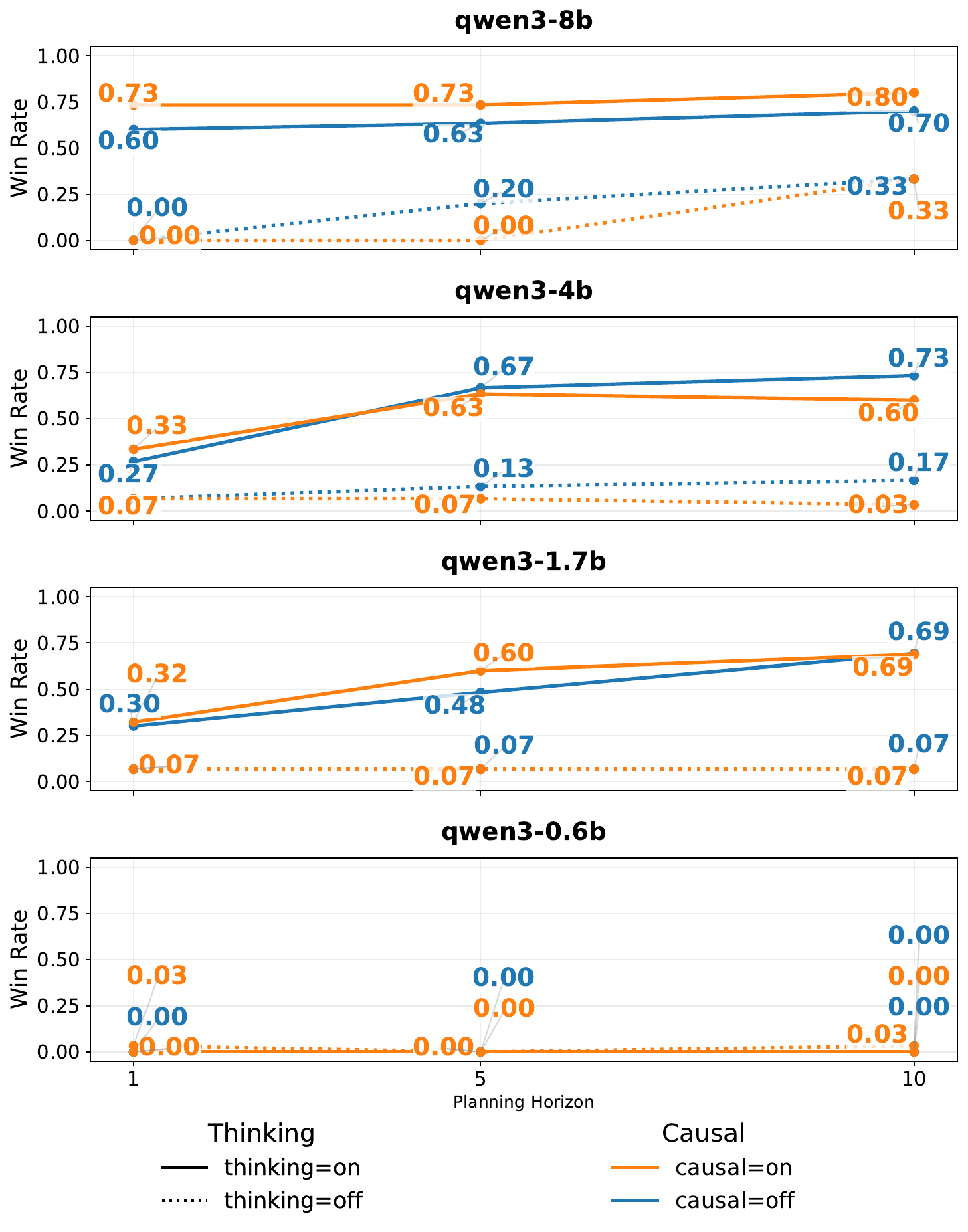}
    \caption{Win-rates vs. planning horizon for the models studied. Lines differentiate the configurations of variables causal augmentation and Thinking (on/off).}
    \label{fig:per_model_win_rate}
\end{figure}

\begin{figure}
    \centering
    \includegraphics[width=0.9\columnwidth]{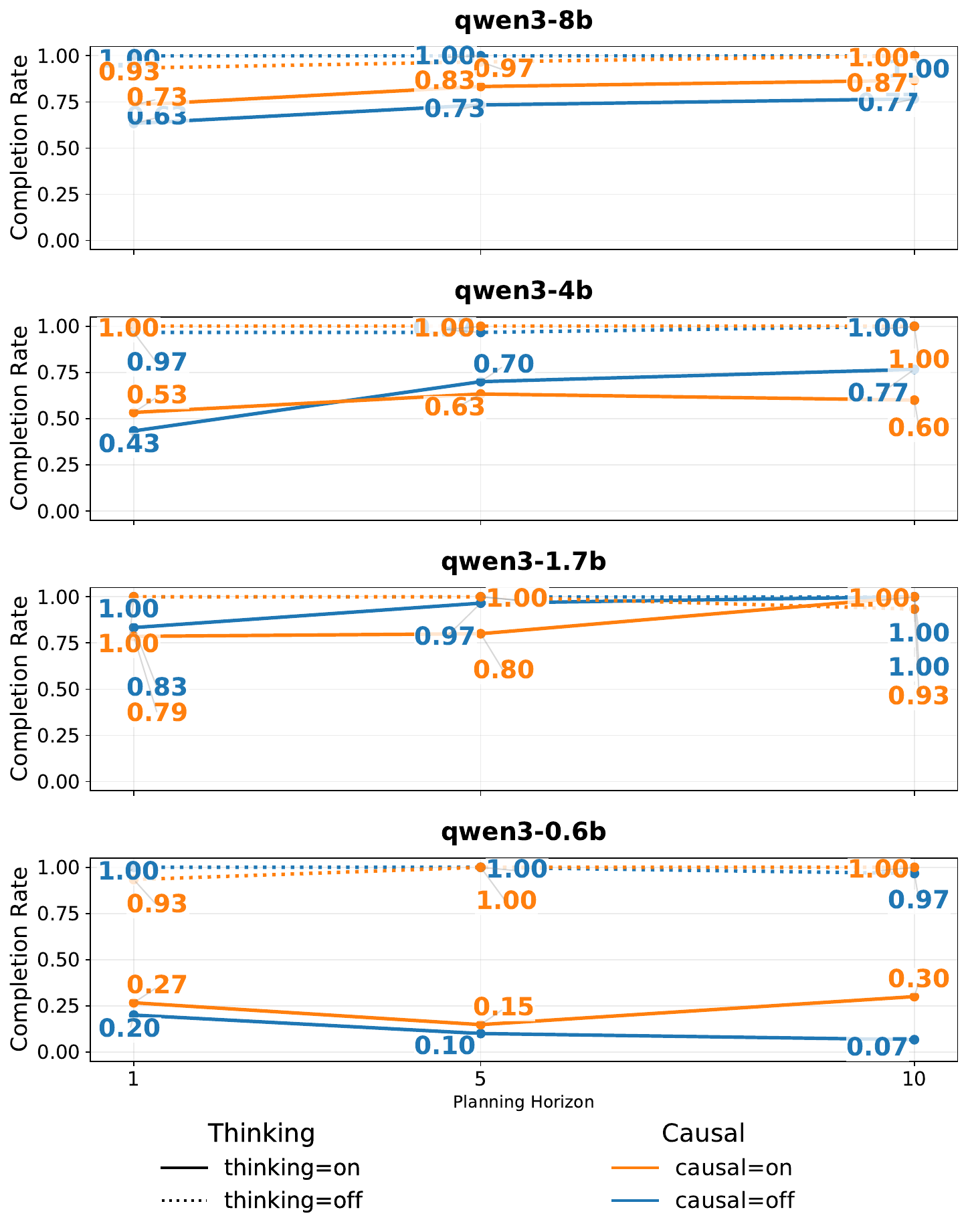}
    \caption{Completion rates vs. planning horizon for the models studied. Lines differentiate the configurations of variables causal augmentation and Thinking (on/off).}
    \label{fig:per_model_completion_rate}
\end{figure}

\begin{figure}
    \centering
    \includegraphics[width=\columnwidth]{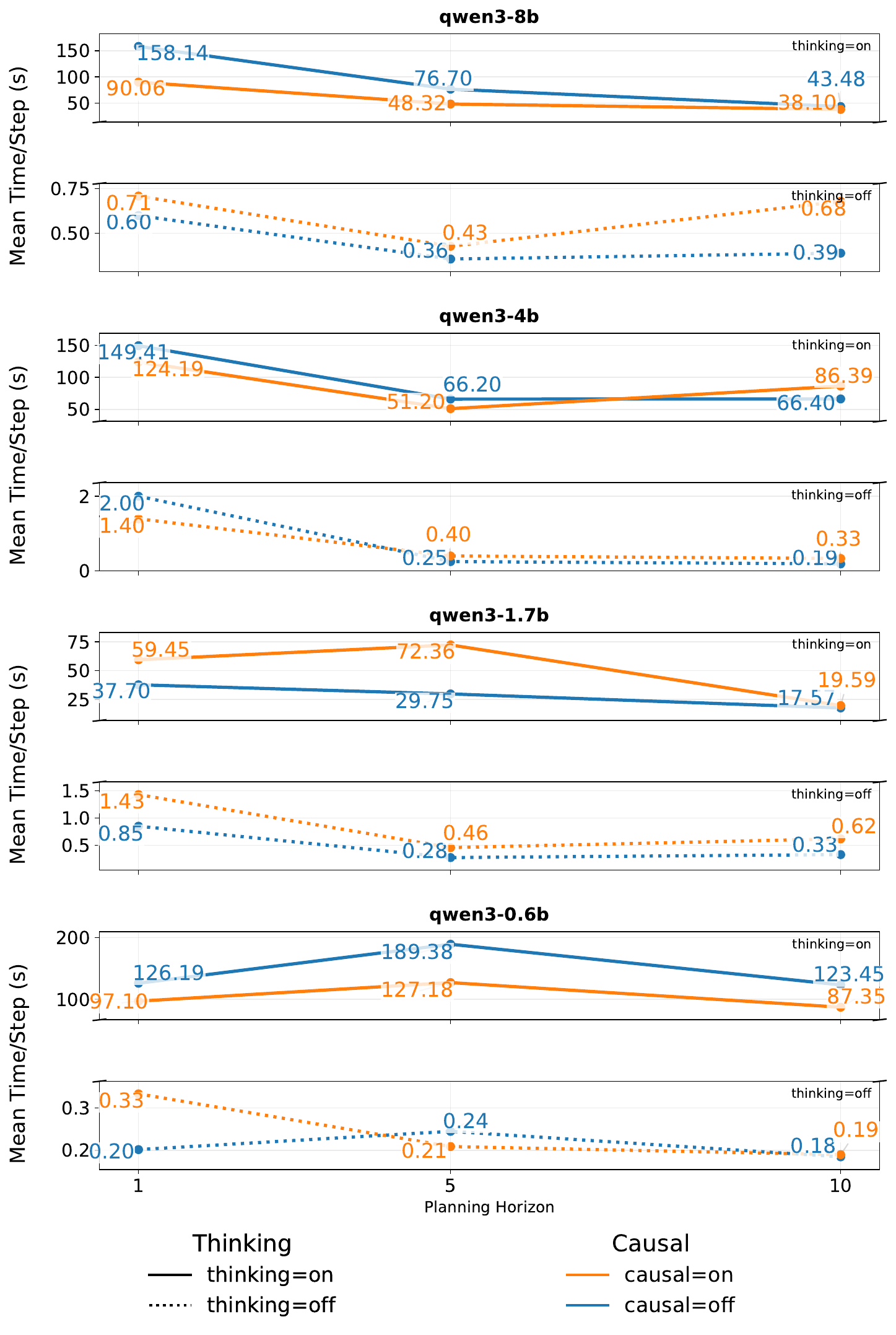}
    \caption{Mean time per step vs. planning horizon for the models studied. Lines differentiate the configurations of variables causal augmentation and Thinking (on/off).}
    \label{fig:per_model_mean_time}
\end{figure}

\Cref{fig:per_model_win_rate}, \Cref{fig:per_model_completion_rate}, \Cref{fig:per_model_mean_time} and \Cref{tab:spatial_mode_summary} show the win-rates, completion rates, and the mean time per step under different causal, thinking, and planning horizons. 

\begin{itemize}
    \item Causal model augmentation shows a marginal overall improvement in the win-rates, while there's a significant improvement when it comes to completion rate and the mean time per step. The win rate improvement becomes important for the \textit{8B} model; however, the same cannot be said for the smaller models.
    \item Thinking or reasoning improves the performance the most, but also comes with a lower completion rate and a stark increase in the response time. This makes the models unreasonable to be used for game playing in practice in their current state.
    \item All the metrics show improvement for longer planning horizons, making it the most beneficial intervention in our study.
\end{itemize}



A more granular analysis can be found in the supplementary material.


\begin{table}[t]
  \centering
  \caption{Average win rate, completion rate, and mean time per step by causal mode, thinking mode, and horizon (all runs combined).}
  \label{tab:spatial_mode_summary}
  \begin{tabular}{lrrr}
    \toprule
    \textbf{Category} & \textbf{Win Rate} & \textbf{Completion} & \textbf{Mean Time/Step (s)} \\
    \midrule
    \textit{Causal} & & & \\
    off & 0.246 & 0.791 & 46.123 \\
    on & 0.250 & 0.809 & 36.538 \\
    \cmidrule(lr){1-4}
    \textit{Thinking} & & & \\
    off & 0.078 & 0.986 & 0.544 \\
    on & 0.433 & 0.597 & 85.776 \\
    \cmidrule(lr){1-4}
    \textit{Horizon} & & & \\
    1 & 0.178 & 0.766 & 53.084 \\
    5 & 0.262 & 0.807 & 40.292 \\
    10 & 0.310 & 0.829 & 29.864 \\
    \bottomrule
  \end{tabular}
\end{table}





\section{Discussion}
\label{sec:discussion}
The LLM-based agents exhibited a strong goal-oriented heuristic, particularly evident in \texttt{spatialgame3}, in which they frequently navigated directly to the terminal goal tile while ignoring intermediate objectives, such as collecting colored keys and unlocking doors. This behavior suggests that while the agents can identify the primary win condition, they struggle with multi-objective optimization and maximizing cumulative rewards within a single trajectory.

Regarding the self-localization experiments, the models demonstrated limited ability to output their exact coordinates. Interestingly, this poor global positioning did not universally translate to failure in gameplay. This discrepancy implies that agents may rely on local, relative spatial reasoning to navigate successfully, rather than maintaining a perfect global state representation. However, this deficiency in absolute spatial grounding is a likely contributor to the imperfect win rates and trajectory inefficiencies observed both in our benchmark and in prior literature, such as the struggles noted by Li et al.~\cite{li2025gvgai_llm}.

A critical operational challenge in deploying LLMs as game agents is the necessity for strict output formatting to enable executable environment actions. Extracting parseable actions from the reasoning trace requires a high volume of heuristic parsing decisions. In instances where model outputs were unparseable, we opted to re-prompt the model rather than default to a fallback action, as automated fallbacks would obfuscate the model's true behavioral capabilities and contaminate the analysis. This parsing fragility often stems from models becoming trapped in repetitive reasoning loops, a known limitation of smaller models~\cite{pipis2025wait}. Ultimately, these formatting bottlenecks underscore that prompt strategy remains the primary tuning factor for stabilizing sequential decision-making in LLMs.

\section{Limitations}
\label{sec:limitations}
We discuss limitations of this work in alignment with the guidelines of Wohlin et al.~\cite{wohlin2012experimentation}.
\vspace{0.5em}
\paragraph{Construct validity} 
Because the benchmark is embedded in a playable game, it is inherently entangled with additional game-specific components. The developed tasks combine spatial formatting, symbolic state representation and interpretation, and action formatting. Nevertheless, their customization is intended to isolate spatial reasoning as far as possible.
\vspace{0.5em}
\paragraph{Internal validity} 
The current study treats causal support only as a prompt context rather than as an integrated causal architecture. 
While this choice is motivated by related work~\cite{jiwatode2026trace2mechanics}, it may still result in the LLM-based agents ignoring the added information. 
In terms of planning horizon, we employed fixed short horizons rather than an adaptive replanning policy. 
However, given the deterministic design of the environment, fixed horizons were sufficient to capture the relevant decision structure while keeping the evaluation controlled and interpretable.
\vspace{0.5em}
\paragraph{Conclusion validity} 
The small benchmark may limit the statistical breadth of our findings; however, this is an intentional feature of the work, as it enables a focused evaluation of the agent's spatial reasoning capabilities.
\vspace{0.5em}
\paragraph{External validity} 
The presented experiments study the capabilities of four LLM models due to practical local-compute constraints. Results may change for smaller/larger models, different models of similar size but different architecture, or a different prompt design.
The benchmark simplifies many elements that make broader GVGAI tasks difficult, including moving enemies, hazards, or other complex dynamics between objects. Therefore, the present benchmark focuses on one capability and does not replace broader evaluation of LLM capabilities in games or other interactive environments.
However, these simplifications isolate the spatial problem and are hence appropriate for this work.


\section{Conclusion and Future Work}
\label{sec:conclusion}
In this work, we tested LLM-based game agents in the context of GVGAI to analyze LLMs' shortcomings and evaluate approaches to tackle those.
To evaluate the agent's performance, we introduce a spatial benchmark inside GVGAI, holding three custom games with five difficulty levels each. 
This benchmark isolates symbolic navigation over tasks of varying difficulty. 
We further analyze how the augmentation of a structured causal model to the agent's prompt affected its performance in the spatial reasoning games.
This showed that on average for Qwen3-8B, causal models relatively improved the win rates by $\approx$~\fpeval{round(\diffcausalwin,2)}\% and reduced the response time per step by $\approx$~\fpeval{round(\diffcausaltime,0)} secs.
Furthermore, multi-step planning resulted in an improvement of up to 45\% while also significantly reducing the response time per step.

Future work should build directly on the limitations identified in this study. Because our current setup uses fixed short planning horizons and represents causal knowledge only as prompt context, the next step is to move toward adaptive replanning, more reliable action interfaces, and causal support integrated into the decision loop itself. This is especially important in more dynamic GVGAI settings than those covered by our current spatial games, where moving enemies, hazards, or blocked routes can invalidate an open-loop plan. In such settings, the planner should output not only intended next actions but also explicitly abort conditions for early re-planning. This would shift the agent from brittle open-loop execution to a guarded planning policy that can replan before disruption becomes failure. Finally, shifting from external VGDL-based causal descriptions to an online induction framework would allow the agent to adapt its causal knowledge dynamically as game rules or environmental hazards are encountered.

\subsection*{Acknowledgments} 
Manuscript revisions for readability and flow were facilitated by Gemini AI. Separately, ChatGPT Codex was employed to support the technical implementation and coding of the study's experiments.

\bibliographystyle{IEEEtran}
\bibliography{references}

\clearpage

\appendices

\section{Prompt to translate VGDL into natural language and an example of generated natural language description}
\begin{figure}[htbp]
  \centering
  \includegraphics[width=\columnwidth]{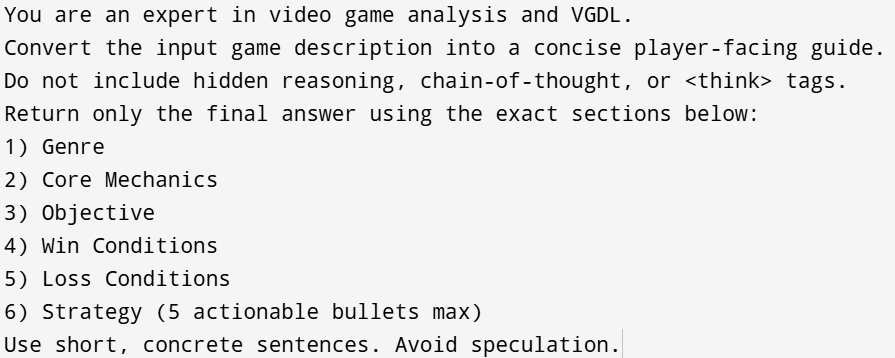}
  \caption{Prompt used to translate VGDL into natural language}
  \label{fig:prompt_example}
\end{figure}

\begin{figure}[htbp]
  \centering
  \includegraphics[width=\columnwidth]{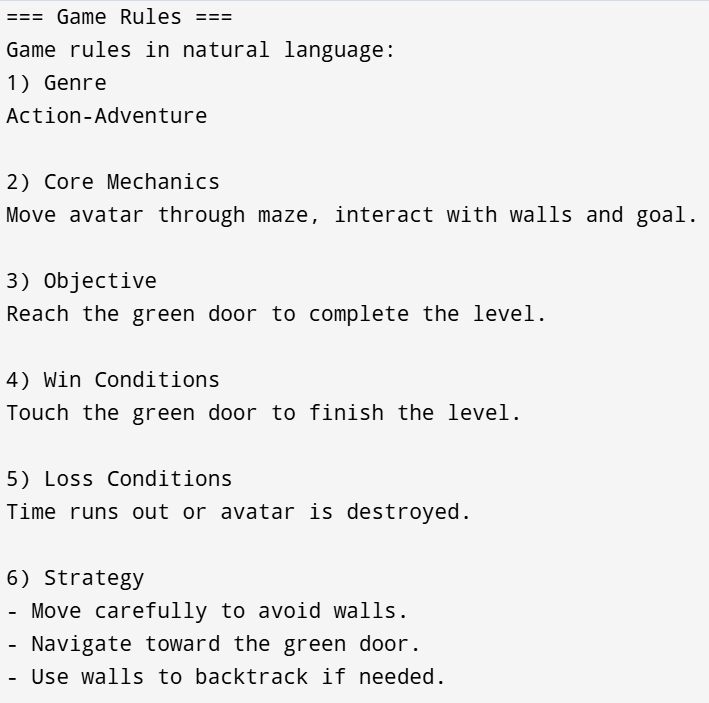}
  \caption{An example of the natural language description of the game generated by the LLM from VGDL code.}
  \label{fig:description_example}
\end{figure}

\newpage
\section{SCM blueprint}
\begin{figure}[htbp]
\begin{lstlisting}[language=Python, caption={SCM Blueprint Definition}, label={lst:scm_blueprint}]
nodes={
    # Design Layer
    "EntityTypes": "Static sprite types, roles (avatar, wall), & attrs (speed, hp). Not time-varying.",
    "ActionSpace": "Agent interventions/actions available at each step (move, shoot).",
    "GlobalMechanics": "Global state transitions independent of collisions (gravity, timers).",
    "InteractionMechanics": "Collision rules mapping (Entities, State_t, Action_t) -> State_{t+1} events (spawn, destroy).",
    "RewardMechanics": "Maps State_t/Interactions to Reward_t or Score_{t+1}.",
    "TerminationMechanics": "Win/loss conditions based on StateVariables (e.g., timeout, goals_met).",
    # Dynamics Layer
    "StateVariables": "Dynamic vars evolving over t (pos, hp). Eq: X_{t+1}=f(PA_X_t, Action_t, U_X).",
    "InitialState": "StateVariables at t=0 derived from EntityTypes + LevelEncoding.",
    # Observation Layer
    "LevelEncoding": "ASCII grid mapping chars to entity instances/positions; induces InitialState.",
},
edges=[
    ("EntityTypes", "InitialState"), ("LevelEncoding", "InitialState"), ("EntityTypes", "StateVariables"),
    ("ActionSpace", "InteractionMechanics"), ("ActionSpace", "StateVariables"),
    ("GlobalMechanics", "StateVariables"), ("InteractionMechanics", "StateVariables"),
    ("StateVariables", "InteractionMechanics"), ("StateVariables", "RewardMechanics"),
    ("InteractionMechanics", "RewardMechanics"), ("RewardMechanics", "StateVariables"),
    ("StateVariables", "TerminationMechanics"), ("InteractionMechanics", "TerminationMechanics"),
],
scm_notes=[
    "Define DYNAMIC SCM with discrete time steps t=0,1...",
    "Separate STATIC design vars from DYNAMIC state vars.",
    "For dynamic X, define X_{t+1} = f(Parents_t, Action_t, Noise).",
    "Define Reward_t & Termination explicitly as equations.",
    "Output machine-readable format (JSON) with: static_nodes, dynamic_variables, edges, equations.",
    "Link LevelEncoding chars to InitialState.",
],
\end{lstlisting}
\end{figure}

\newpage
\clearpage
\section{VGDLs for the games and level layouts}

\subsection{spatialgame1}
\begin{figure}[h] 
\centering
\begin{minipage}{0.48\textwidth}
\begin{lstlisting}[numbers=none, caption={Levels L0, L1, and L2 for SpatialGame1}]
L0          L1           L2
wwwwww      wwwwwww      wwwwwwww
wA...w      wA....w      wA..w..w
w.www.      w.www.w      w.w.w..w
w...gw      w...w.w      w.w....w
w....w      w.w...w      w...ww.w
wwwwww      w...g.w      w.w....w
            wwwwwww      w...wg.w
                         wwwwwwww
\end{lstlisting}
\end{minipage}
\hfill
\begin{minipage}{0.48\textwidth}
\begin{lstlisting}[numbers=none, caption={Levels L3 and L4 for SpatialGame1}]
L3             L4
wwwwwwwww      wwwwwwwwww
wA..w...w      wA..w....w
w.w.w.w.w      w.w.w.ww.w
w.w...w.w      w.w...w..w
w.www.w.w      w.www.w..w
w...w...w      w...w.w..w
w.w.www.w      www.w.w..w
w....g..w      w...w....w
wwwwwwwww      w.wwwwwg.w
               wwwwwwwwww
\end{lstlisting}
\end{minipage}

\vspace{1em}

\begin{minipage}{0.98\columnwidth}
\begin{lstlisting}[caption={VGDL Game Description for SpatialGame1}, label={lst:vgdl}]
BasicGame key_handler=Pulse square_size=40
    SpriteSet
        floor > Immovable hidden=True img=oryx/floor3
        avatar > MovingAvatar img=oryx/knight1
        goal > Door color=GREEN img=oryx/doorclosed1
        wall > Immovable img=oryx/wall3 autotiling=True

    LevelMapping
        A > avatar floor
        g > goal floor
        w > wall floor
        . > floor

    InteractionSet
        avatar wall > stepBack
        goal avatar > killSprite scoreChange=1

    TerminationSet
        SpriteCounter stype=goal limit=0 win=True
        SpriteCounter stype=avatar limit=0 win=False
        Timeout limit=6000 win=False
\end{lstlisting}
\end{minipage}
\end{figure}

\vspace{100pt}

\subsection{spatialgame2}

\begin{figure}[h] 
\centering
\begin{minipage}{0.48\textwidth}
\begin{lstlisting}[numbers=none, caption={Levels L0, L1, and L2 for SpatialGame2}]
L0          L1           L2
wwwwww      wwwwwww      wwwwwwww
wA.kew      wA..w.w      wA...w.w
w....w      w.w.k.w      w.w.wk.w
w....w      w.w...w      w.w.w..w
w....w      w...w.w      w...ww.w
wwwwww      w..e..w      w.w....w
            wwwwwww      w...e..w
                         wwwwwwww
\end{lstlisting}
\end{minipage}
\hfill
\begin{minipage}{0.48\textwidth}
\begin{lstlisting}[numbers=none, caption={Levels L3 and L4 for SpatialGame2}]
L3             L4
wwwwwwwww      wwwwwwwwww
wA..w...w      wA..w....w
w.w.w.w.w      w.w.w.ww.w
w.w.k.w.w      w.w.k.w..w
w.www.w.w      w.www.w..w
w...w...w      w...w.w..w
w.w.www.w      www.w.w..w
w....e..w      w...w....w
wwwwwwwww      w.wwwwwe.w
               wwwwwwwwww
\end{lstlisting}
\end{minipage}

\vspace{1em}

\begin{minipage}{0.98\columnwidth}
\begin{lstlisting}[caption={VGDL Game Description for SpatialGame2}, label={lst:vgdl2}]
BasicGame key_handler=Pulse square_size=40
    SpriteSet
        floor > Immovable hidden=True img=oryx/floor3
        avatar > MovingAvatar
            nokey > img=oryx/rogue
            withkey > color=ORANGE img=oryx/swordmankey1
        key > Passive img=oryx/key2
        exit > Door color=GREEN img=oryx/doorclosed1
        wall > Immovable img=oryx/wall3 autotiling=True

    LevelMapping
        A > nokey floor
        k > key floor
        e > exit floor
        w > wall floor
        . > floor

    InteractionSet
        avatar wall > stepBack
        nokey key > transformTo stype=withkey scoreChange=1 killSecond=True
        exit nokey > stepBack
        exit withkey > killSprite scoreChange=2

    TerminationSet
        SpriteCounter stype=exit limit=0 win=True
        SpriteCounter stype=avatar limit=0 win=False
        Timeout limit=6000 win=False
\end{lstlisting}
\end{minipage}
\end{figure}

\vspace{50pt}

\subsection{spatialgame3}

\begin{figure}[htbp] 
\centering
\begin{minipage}{0.48\textwidth}
\begin{lstlisting}[numbers=none, caption={Levels L0, L1, and L2 for SpatialGame3}]
L0          L1           L2
wwwwww      wwwwwww      wwwwwwww
wA.r1w      wA..r1w      wA..w..w
w....w      w.w.w.w      w.wrw1.w
w..b2w      w...b2w      w.w.w..w
w...gw      w.w...w      w..b.2.w
wwwwww      w...g.w      w.w....w
            wwwwwww      w...g..w
                         wwwwwwww
\end{lstlisting}
\end{minipage}
\hfill
\begin{minipage}{0.48\textwidth}
\begin{lstlisting}[numbers=none, caption={Levels L3 and L4 for SpatialGame3}]
L3             L4
wwwwwwwww      wwwwwwwwww
wA..w...w      wA..w....w
w.wrw1w.w      w.wrw1ww.w
w.w...w.w      w.w...w..w
w.www.w.w      w.www.w..w
w..b.2..w      w..b.2w..w
w.w.www.w      www.w.w..w
w....g..w      w...w....w
wwwwwwwww      w.wwwwwg.w
               wwwwwwwwww
\end{lstlisting}
\end{minipage}

\vspace{1em}
\end{figure}

\begin{figure}[t] 
\centering

\begin{minipage}{0.98\columnwidth}
\begin{lstlisting}[caption={VGDL Game Description for SpatialGame3}, label={lst:vgdl3}]
BasicGame key_handler=Pulse square_size=40
    SpriteSet
        floor > Immovable hidden=True img=oryx/floor3
        avatar > MovingAvatar img=oryx/rogue

        key > Resource limit=1
            redkey > img=oryx/potion3
            bluekey > img=oryx/potion1

        door > Immovable
            reddoor > img=oryx/slime3
            bluedoor > img=oryx/slime1

        goal > Door color=GREEN img=oryx/doorclosed1
        wall > Immovable img=oryx/wall3 autotiling=True

    LevelMapping
        A > avatar floor
        r > redkey floor
        b > bluekey floor
        1 > reddoor floor
        2 > bluedoor floor
        g > goal floor
        w > wall floor
        . > floor

    InteractionSet
        avatar wall door > stepBack
        key avatar > collectResource scoreChange=1

        reddoor avatar > killIfOtherHasMore resource=redkey limit=0 scoreChange=1
        avatar reddoor > changeResource resource=redkey value=-1

        bluedoor avatar > killIfOtherHasMore resource=bluekey limit=0 scoreChange=1
        avatar bluedoor > changeResource resource=bluekey value=-1

        goal avatar > killSprite scoreChange=3

    TerminationSet
        SpriteCounter stype=goal limit=0 win=True
        SpriteCounter stype=avatar limit=0 win=False
        Timeout limit=6000 win=False
\end{lstlisting}
\end{minipage}
\end{figure}

\newpage
\clearpage

\section{Extended Spatial Reasoning Visualizations}

\subsection{Causal \& thinking diagnostics}
This single causal heatmap highlights which combinations of causal context and horizon most consistently increase win rates, supporting the planning-enabled arguments in the main text.

\begin{figure}[htbp]
  \centering
  \includegraphics[width=\columnwidth]{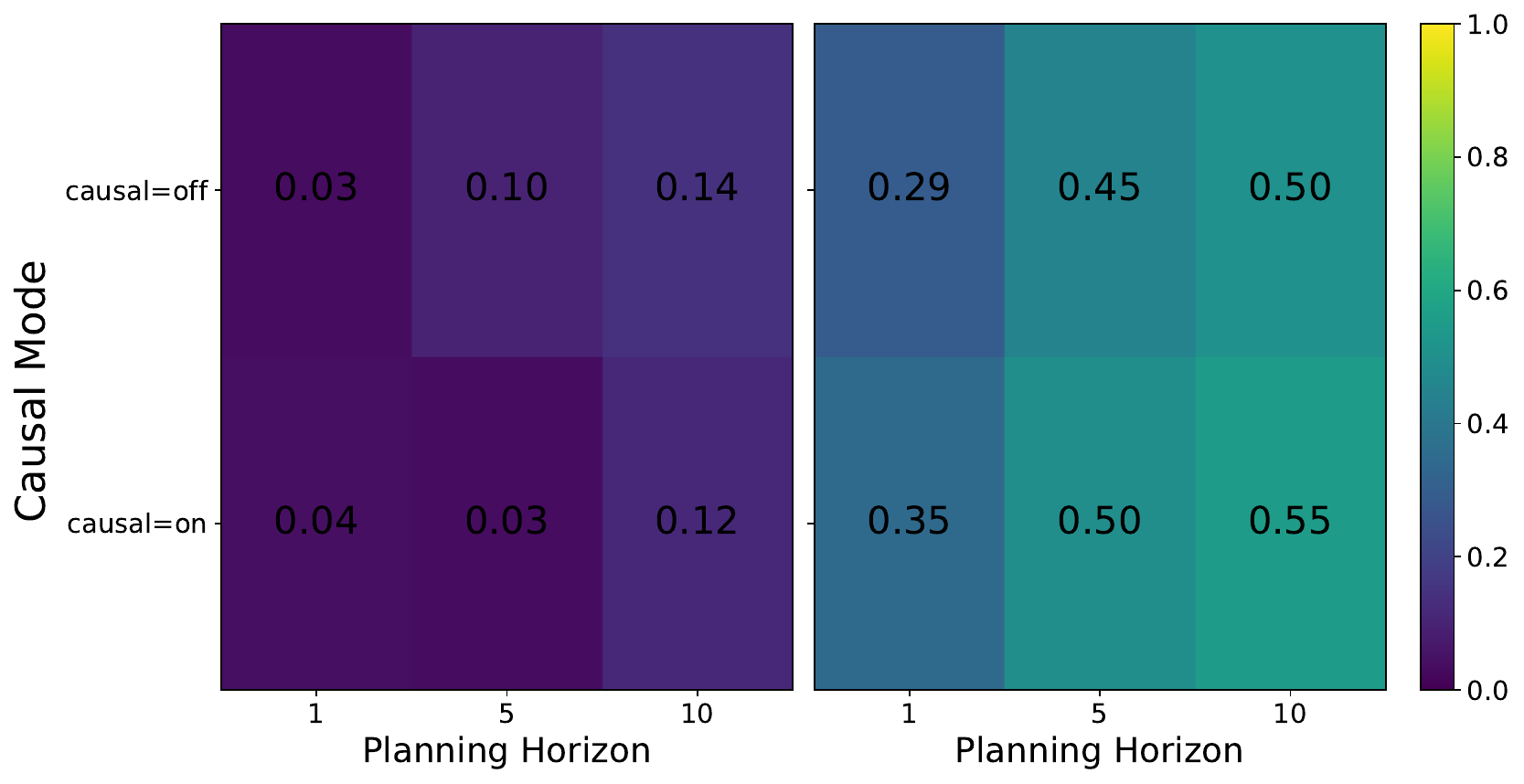}
  \caption{Win-rate heatmap comparing causal vs non-causal prompts across horizons. The heatmap on the left is for thinking=off and on the right for thinking=on}
\end{figure}

\newpage
\clearpage

\subsection{Per-game, per-model, and per-level breakdowns}
These plots zoom into how different models behave on the benchmarked games, tracking completion, timing, and win-percentage statistics.

\begin{figure}[htbp]
  \centering
  \includegraphics[width=\columnwidth]{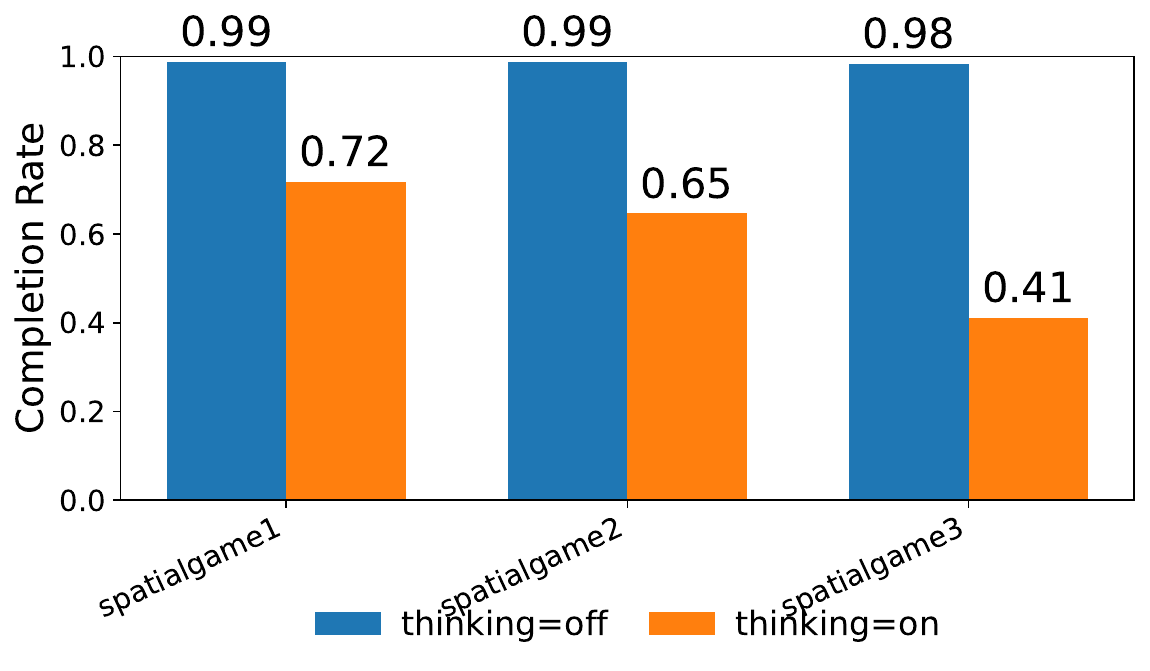}
  \caption{Completion rate per game when thinking mode is toggled.}
\end{figure}

\begin{figure}[htbp]
  \centering
  \includegraphics[width=\columnwidth]{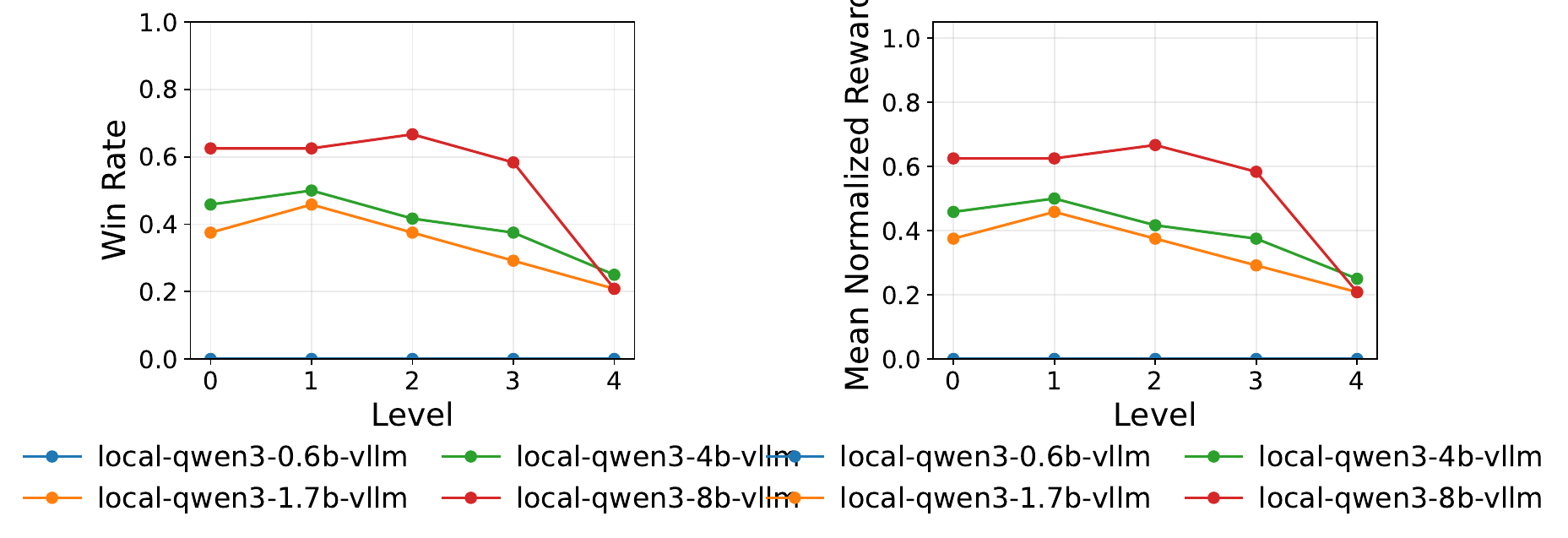}
  \caption{Per-model completion for SpatialGame1.}
\end{figure}

\begin{figure}[htbp]
  \centering
  \includegraphics[width=\columnwidth]{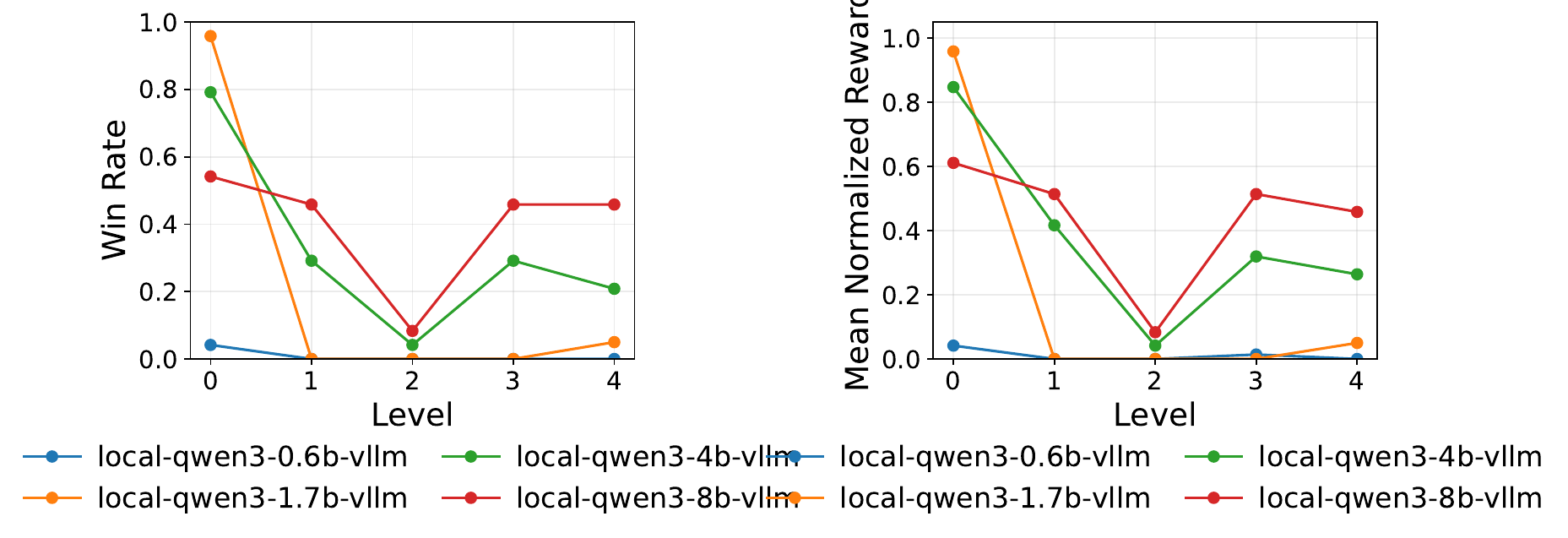}
  \caption{Per-model completion for SpatialGame2.}
\end{figure}

\begin{figure}[htbp]
  \centering
  \includegraphics[width=\columnwidth]{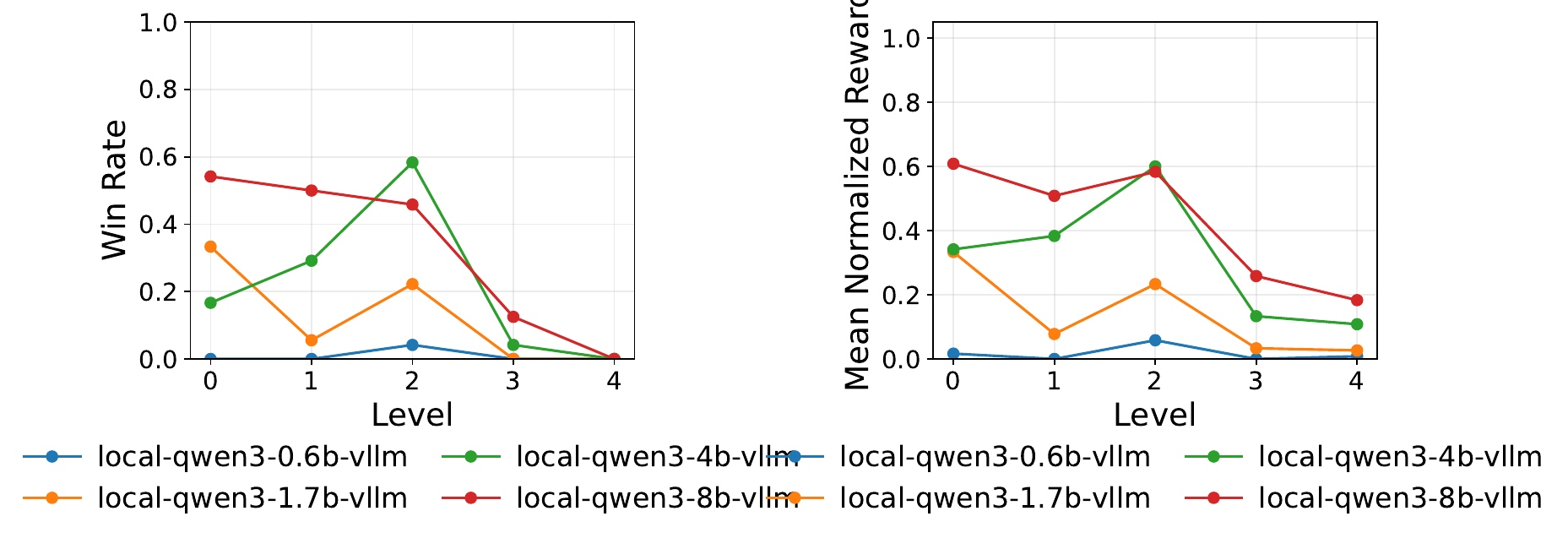}
  \caption{Per-model completion for SpatialGame3.}
\end{figure}

\begin{figure}[htbp]
  \centering
  \includegraphics[width=\columnwidth]{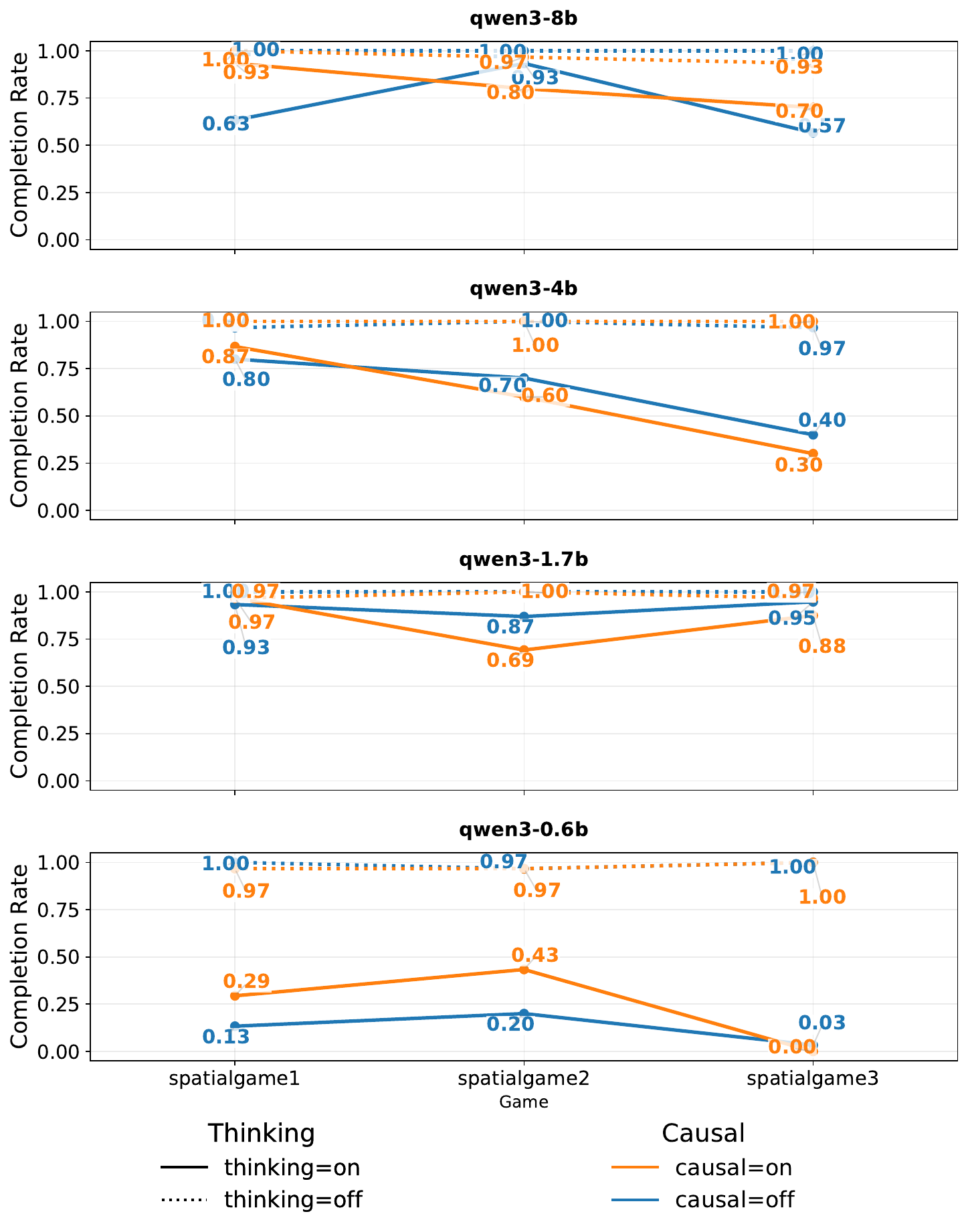}
  \caption{Completion rate per model aggregated across games.}
\end{figure}

\begin{figure}[htbp]
  \centering
  \includegraphics[width=\columnwidth]{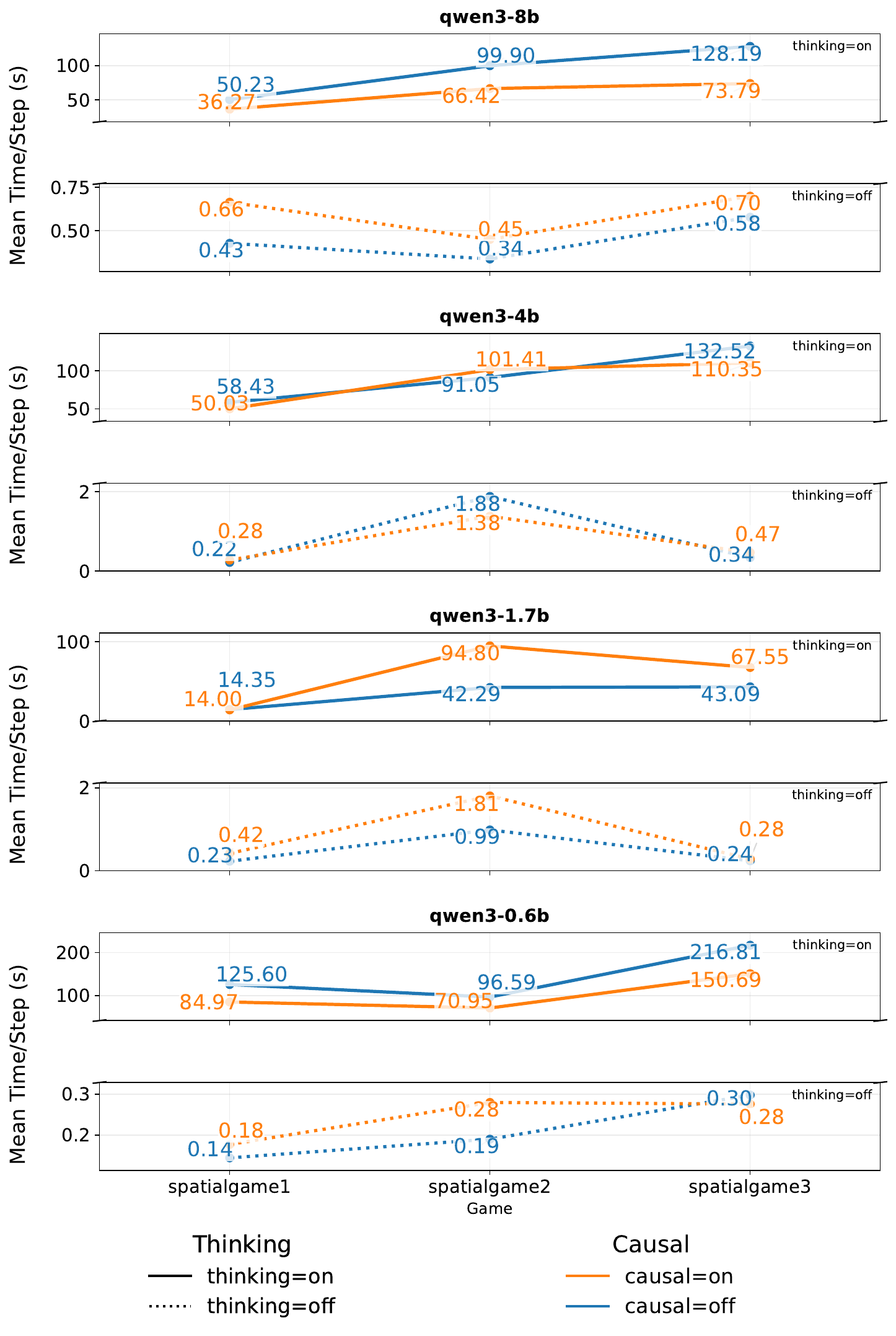}
  \caption{Mean time per step grouped by game for each model.}
\end{figure}

\begin{figure}[htbp]
  \centering
  \includegraphics[width=\columnwidth]{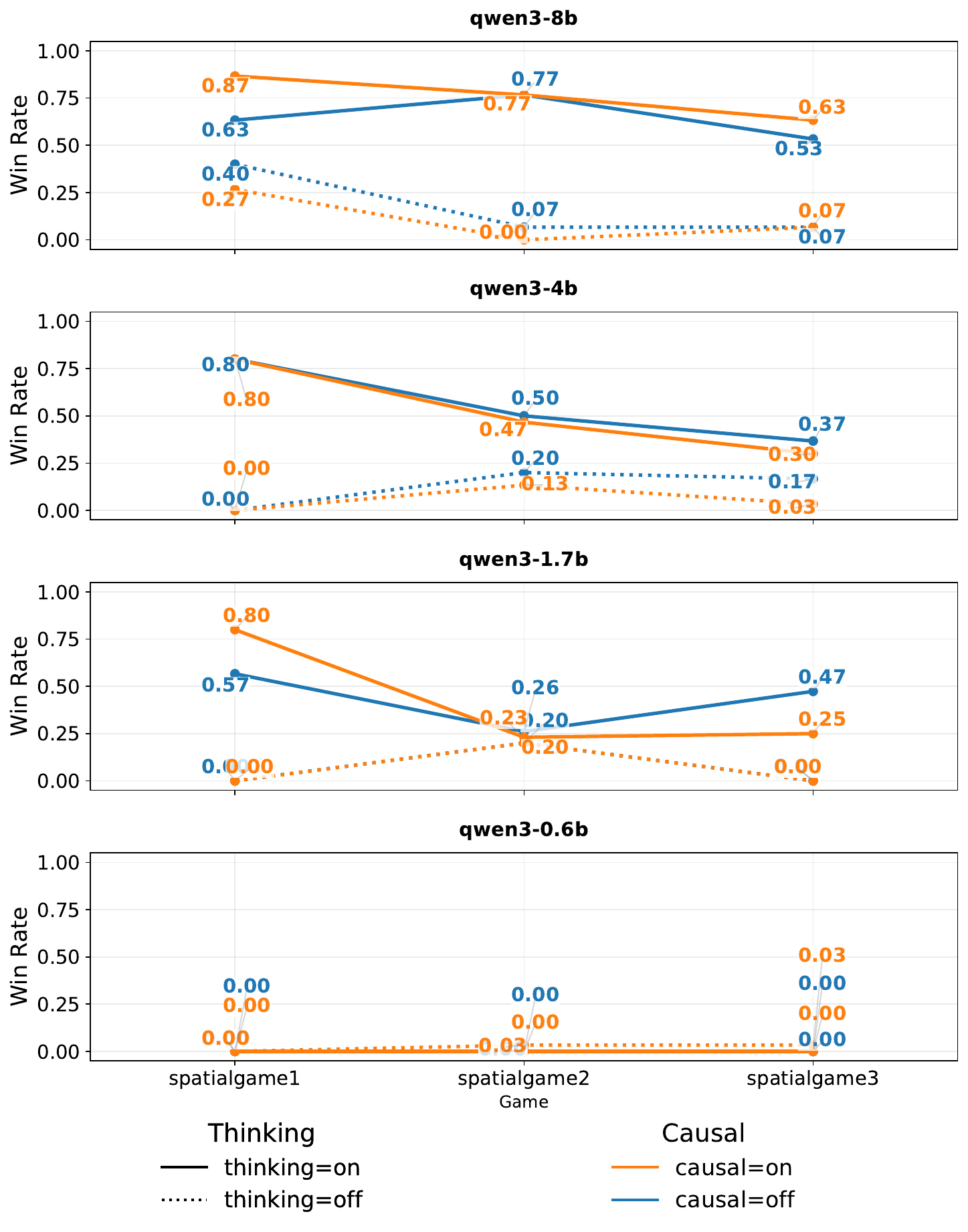}
  \caption{Win rate per game for each model.}
\end{figure}

\newpage
\clearpage

\subsection{Heatmaps \& distribution snapshots}
These dense snapshots show where completion and win-rate failures cluster across game/level and model pairings.

\begin{figure}[htbp]
  \centering
  \includegraphics[width=\columnwidth]{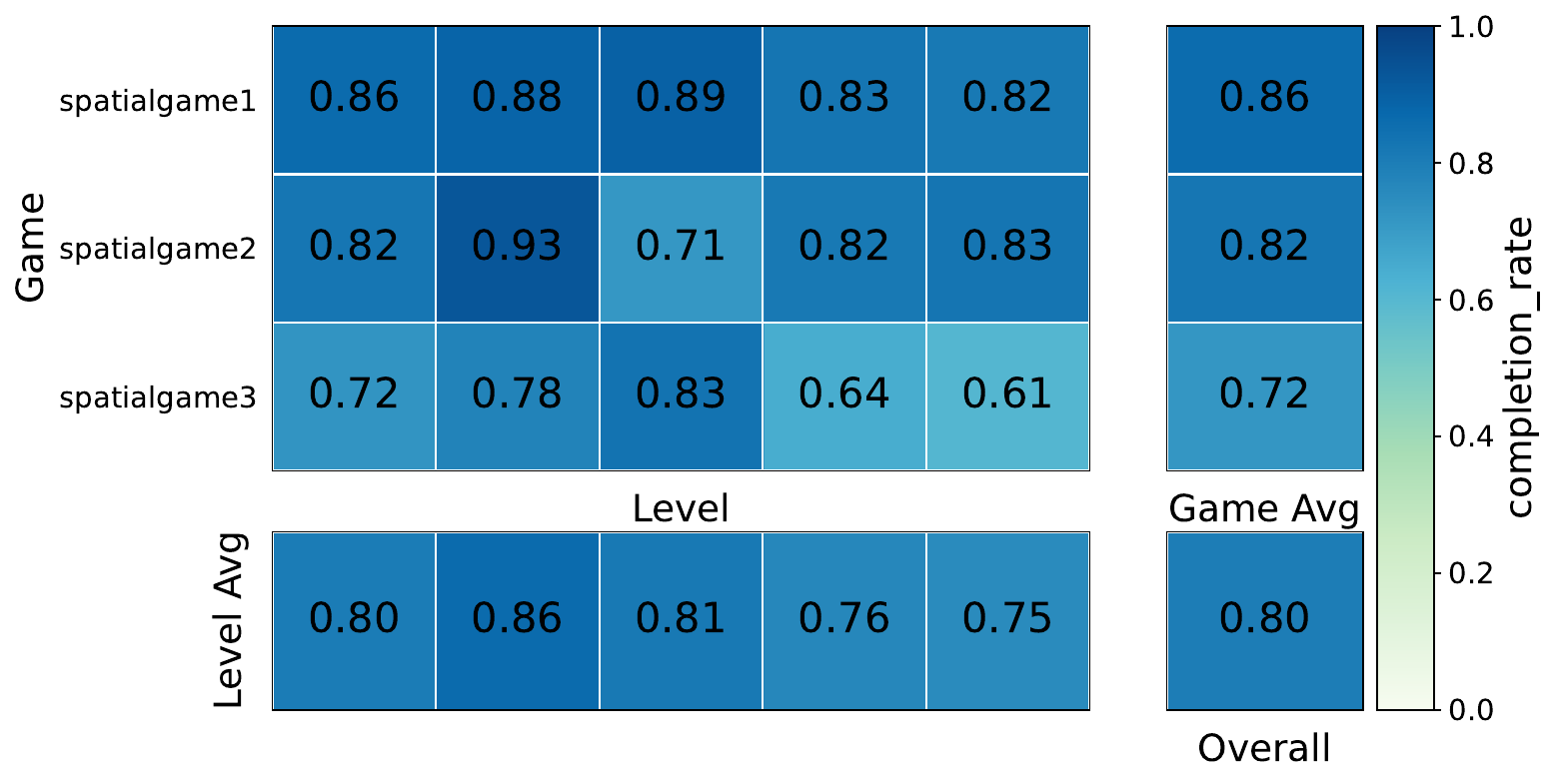}
  \caption{Completion-rate heatmap arranged by game and level difficulty.}
\end{figure}

\begin{figure}[htbp]
  \centering
  \includegraphics[width=\columnwidth]{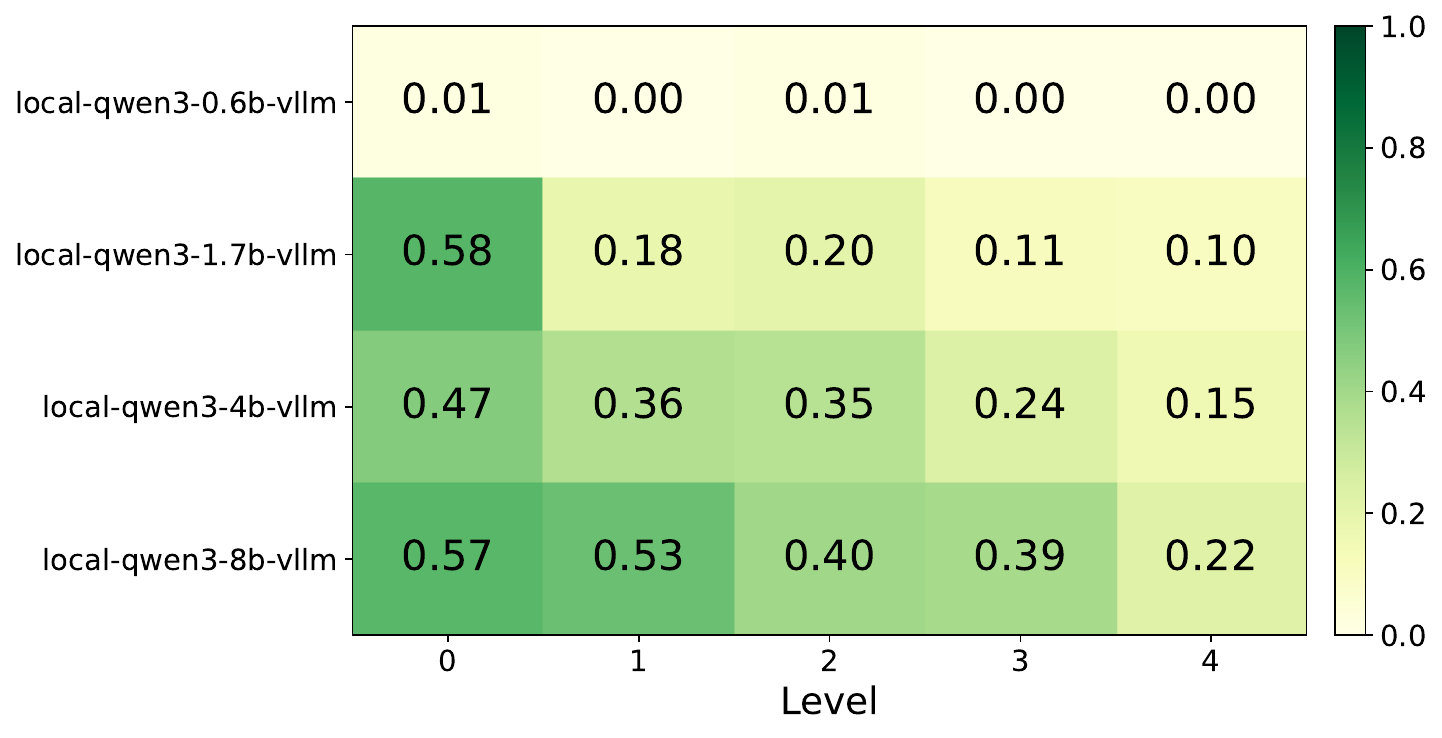}
  \caption{Win-rate heatmap arranged by model scale and level difficulty.}
\end{figure}

\subsection{Summary tables}
The tables below summarize the causal, horizon, and thinking analyses referenced throughout the paper.

\begin{figure*}[htbp]
  \centering
  \includegraphics[width=\textwidth]{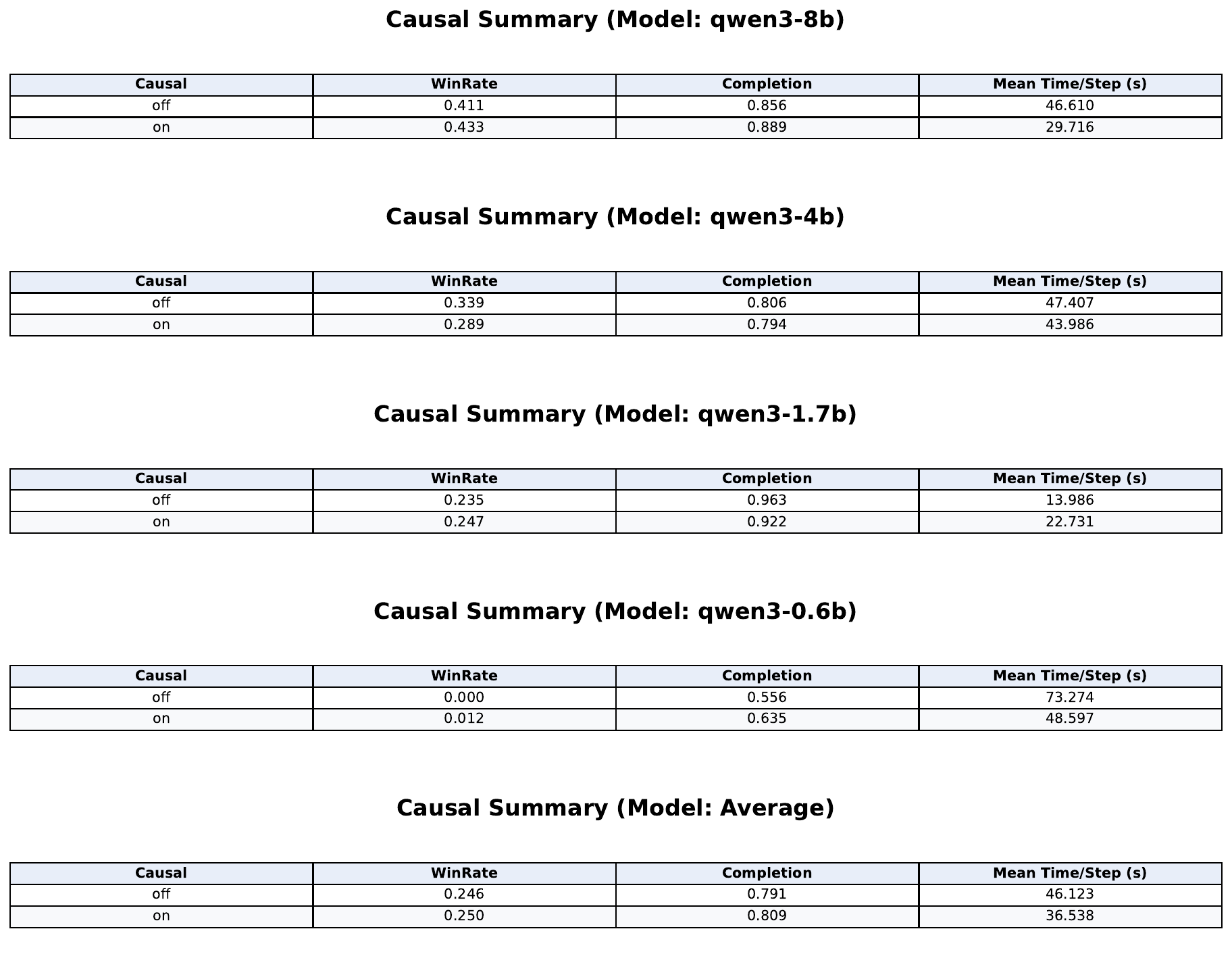}
  \caption{Summary table for the causal prompt experiments.}
\end{figure*}

\begin{figure*}[htbp]
  \centering
  \includegraphics[width=\textwidth]{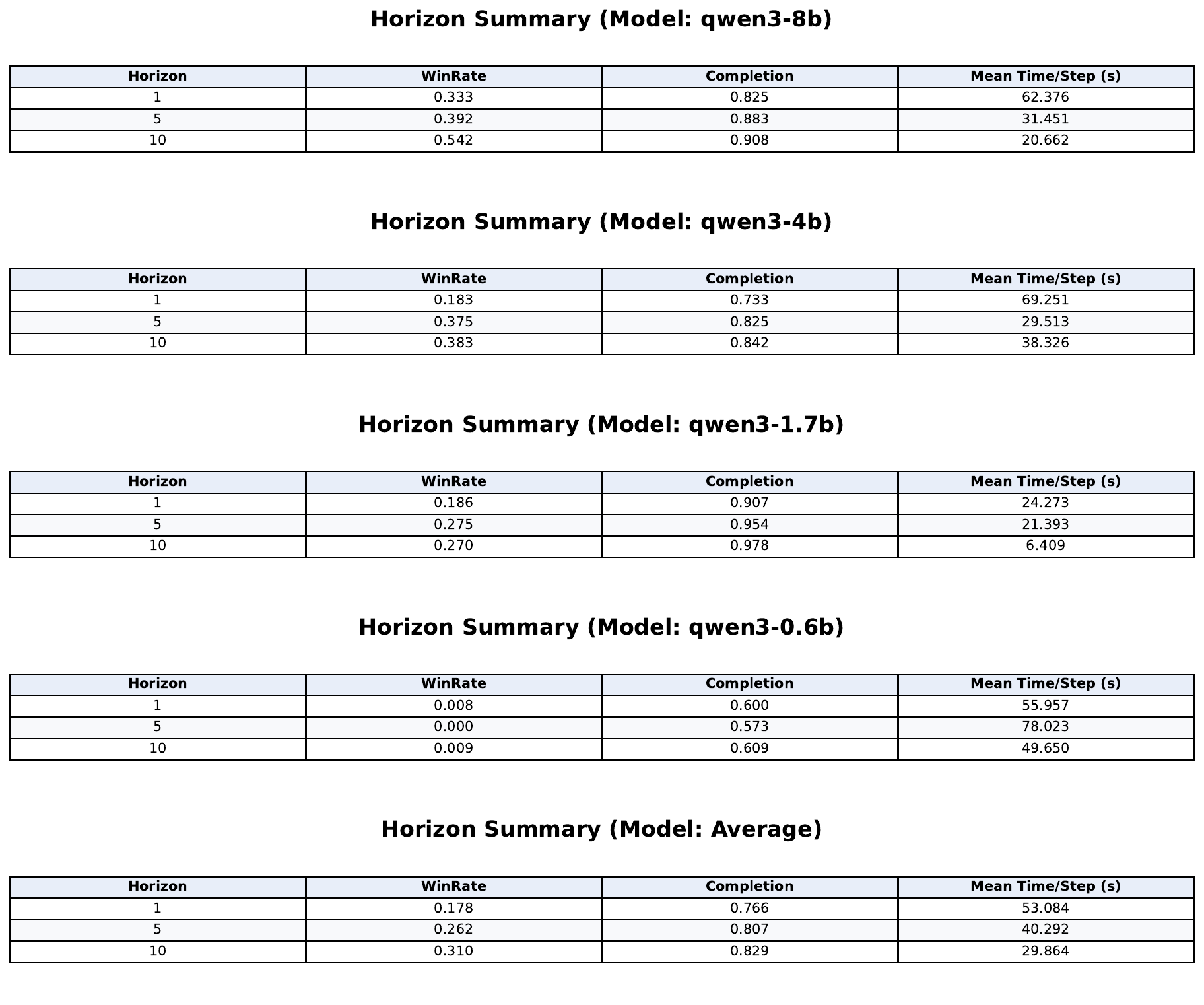}
  \caption{Summary table for the planning horizon comparisons.}
\end{figure*}

\begin{figure*}[htbp]
  \centering
  \includegraphics[width=\textwidth]{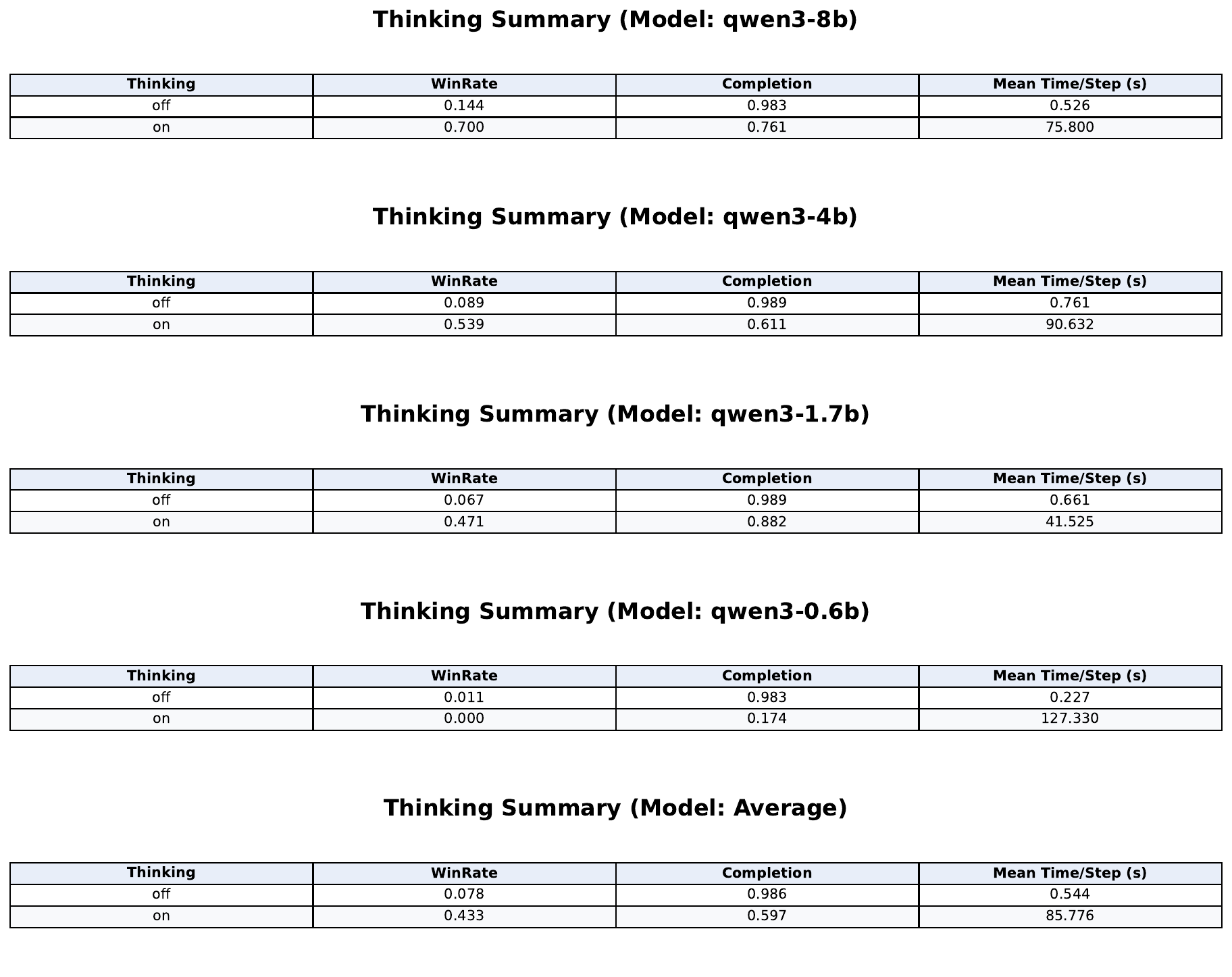}
  \caption{Summary table for the thinking-mode experiments.}
\end{figure*}

\end{document}